\documentclass[11pt,letterpaper]{article}

\usepackage[margin=1in]{geometry}
\usepackage{times}
\usepackage{natbib}
\setcitestyle{authoryear,round,citesep={;},aysep={,},yysep={;}}

\usepackage{amsmath,amsfonts,bm}

\def\eqref#1{equation~\ref{#1}}

\def\1{\bm{1}}

\DeclareMathAlphabet{\mathsfit}{\encodingdefault}{\sfdefault}{m}{sl}
\SetMathAlphabet{\mathsfit}{bold}{\encodingdefault}{\sfdefault}{bx}{n}

\usepackage{amsmath,amssymb}
\usepackage{hyperref}
\usepackage{url}
\usepackage{graphicx}

\hypersetup{hidelinks}

\title{
Beyond Task Success: A Theory of Representation Reuse
}

\author{
Zibo Meng
\and
Andrew C. M. Austin
\and
Alin Achim
}

\date{}

\begin{document}

\maketitle


\begin{abstract}
A learned representation is useful beyond its source task only if later tasks
can make use of what the encoder has already learned. This matters most when
the new task has limited supervision. Source-task performance, however, gives
no guarantee that the representation itself will remain useful once the
original head is removed.
We study representation reuse with respect to a specified family of future
tasks. For such a family, we define the task-relevant state as the coarsest
deterministic state sufficient for all of its tasks. A larger task family can
require a finer state, and losing a required distinction makes at least one
future task unrecoverable on a target support containing that conflict. Exact
reuse also depends on how the retained information is presented to the
downstream rule. We separate this into structural accessibility, which concerns
whether the required distinctions remain recoverable, and coordinate
consistency, which concerns whether those distinctions can be interpreted by
the same downstream rule across samples.
Both failure modes appear separately in the experiments. Reuse improves as
task-relevant distinctions become more accessible. Reversible
sample-dependent coordinate changes can sharply reduce reuse even when the
representation remains recoverable, and inverting the applicable coordinate
change restores performance. The same accessibility--reuse ordering appears
in prospectively specified CIFAR-100 experiments with both ResNet-18 and
DenseNet-121. Finally, preserving a candidate state that contains the
distinctions needed by the anticipated task family improves reuse on compatible
held-out tasks, with the clearest benefit when target supervision is limited.
Representation reuse is therefore relative to the tasks that follow source
learning: a useful representation must retain the distinctions those tasks
need and make them available in a form that a shared downstream rule can use.
\end{abstract}


\section{Introduction}

A learned representation may be reused after its original task changes. Once
the source head is replaced, later tasks depend on distinctions retained and
made usable by the encoder, especially when target supervision is limited.
Source-task performance only shows that the encoder and source head work well
together; it does not reveal which distinctions remain available or can be
interpreted consistently across samples. Two models can therefore match on the
source task yet differ substantially in reuse.

Consider a synthetic-aperture radar (SAR) encoder trained for target
recognition and later reused, with few labeled examples, for change
detection or correspondence. Two encoders can match in classification accuracy
yet expose different information about local geometry, spatial arrangement, or
scattering structure. Source accuracy therefore cannot determine which
representation will support the new task. This motivates our general question:
which distinctions must remain usable when the task changes?

Learned representations are commonly evaluated by frozen probes, feature
transfer, transferability, and sample-efficient adaptation
\citep{alain2017iclr-understanding,NIPS2014_532a2f85,8954384,3524938.3525614,ICLR2024_3209cf33}.
Other work asks what pretraining transfers and how source-induced feature
spaces interact with target tasks and supervision
\citep{NEURIPS2020_0607f4c7,pmlr-v267-tahir25a,ICLR2026_58ab5d6e},
or studies sufficiency, minimality, invariance, equivariance, and
identifiability of learned representations
\citep{8503149,NEURIPS2020_4c2e5eaa,9880437,gupta2024structuring,
pmlr-v108-khemakhem20a,pmlr-v139-roeder21a,ICLR2025_5ee7ed60,ICLR2026_f67e5f99}.
Our question begins after the source decision rule is removed. Different future
tasks can require different distinctions, so reuse cannot in general be
described by a single task-independent score. We therefore define reuse with
respect to a specified family of future tasks.

For a future-task family $\mathcal H$, we define the task-relevant state
$s_{\mathcal H}^{\star}$ as the coarsest deterministic state sufficient for
every task in the family. It captures the distinctions that must remain
available to support all tasks in $\mathcal H$. If
$\mathcal H_1\subseteq\mathcal H_2$, then
$s_{\mathcal H_2}^{\star}$ is at least as fine as
$s_{\mathcal H_1}^{\star}$. Collapsing a distinction required by
$s_{\mathcal H}^{\star}$ makes at least one task in $\mathcal H$
unrecoverable on a target support containing that conflict. Reuse is therefore
defined by the distinctions required by the specified future-task family, not
by the total amount of information retained by the representation.

Recovering the required distinctions is not sufficient for exact reuse. Their
representation must also admit a common downstream interpretation.
\emph{Structural accessibility} concerns whether the required distinctions are
recoverable from the representation. \emph{Coordinate consistency} concerns
whether the same downstream rule can interpret them across samples. A single
globally shared reversible reparameterization can be absorbed by the decoder
under unrestricted decoding. Sample-dependent conventions may prevent a common
decoder even when the representation remains recoverable.

Source-matched models can still differ sharply in reuse. On CIFAR-100, nine
matched representation pairs have a mean absolute source-performance gap of
only $0.0036$, while C1 exceeds C0 by $0.1718$ in low-shot reuse. Reversible
coordinate changes expose a separate failure: reuse falls from $0.4096$ under
a common convention to $0.0154$ under sample-wise conventions even though the
representation remains recoverable; inverting the applicable convention
restores reuse to $0.4096$. In prospectively specified CIFAR-100 experiments,
held-out reuse increases from Q1 through Q5 for both ResNet-18 and
DenseNet-121. On compatible held-out tasks, preserving a state that contains
the distinctions required by the anticipated task family improves low-shot
reuse by $0.1715$; the full-support gain is $0.0241$ and its confidence
interval includes zero. The practical effect is clearest when target supervision is limited, where
the measured reuse advantage is largest.

\noindent\textbf{Contributions.}
We separate representation reuse from source-task success and show that
source-matched representations can differ in reuse and can change ordering
across future-task families. We formalize this dependence by defining the
coarsest deterministic state sufficient for a specified family, deriving
refinement and insufficient-state impossibility results, and characterizing
exact reuse through structural accessibility and coordinate consistency.
Intervention and rescue experiments test these two failure modes, including
reversible coordinate changes that preserve recoverability while reducing
blind reuse. We then connect the theory to source learning: preserving a
candidate state that refines the anticipated task requirements improves
low-shot reuse on compatible held-out tasks, without implying a
task-independent advantage for unrelated families.

\begin{figure}[!ht]
    \centering
    \includegraphics[width=0.90\linewidth]{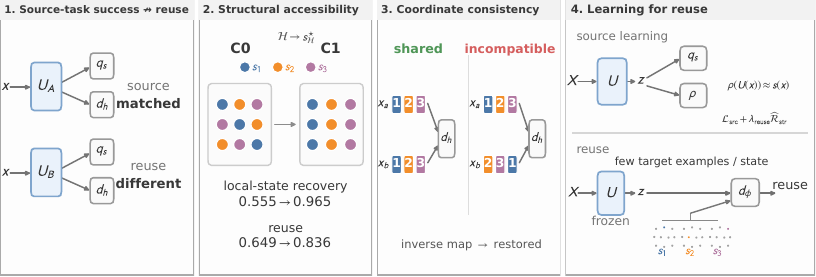}
    \caption{
    \textbf{Representation reuse.}
    \textbf{(1)} Matched source performance need not imply matched reuse.
    \textbf{(2)} Future-task distinctions must remain structurally accessible.
    \textbf{(3)} Retained distinctions must admit a shared interpretation.
    \textbf{(4)} Structure-preserving source learning promotes recoverability
    of an anticipated candidate state.
    }
    \label{fig:overview}
\end{figure}


\section{A Task-Relative Theory of Representation Reuse}
\label{sec:representation_reuse}


\subsection{From Source-Task Learning to Representation Reuse}
\label{sec:reuse_setup}

Let
$U_\theta:\mathcal X\rightarrow\mathcal Z$
be an encoder and
$q_\psi:\mathcal Z\rightarrow\mathcal Y_s$
a source head. Source learning minimizes
\begin{equation}
\mathcal R_{\mathrm{src}}(\theta,\psi)
=
\mathbb E_{(x,y_s)\sim P_s}
\left[
\ell_s
\left(
q_\psi(U_\theta(x)),y_s
\right)
\right].
\label{eq:source_risk}
\end{equation}
Let $(\theta^\ast,\psi^\ast)$ be a minimizer and write
$U=U_{\theta^\ast}$. Low source-task risk concerns the composite predictor
$q_{\psi^\ast}\circ U$ and does not determine which targets remain
recoverable once the source head is removed.

For a target
$h:\mathcal X\rightarrow\mathcal Y_t$,
target distribution $P_t$, and downstream class
$\mathcal D\subseteq\{d:\mathcal Z\rightarrow\mathcal Y_t\}$, define
\begin{equation}
\mathcal R_{\mathrm{reuse}}
\left(
U;h,\mathcal D
\right)
=
\inf_{d\in\mathcal D}
\mathbb E_{x\sim P_t}
\left[
\ell_t
\left(
d(U(x)),h(x)
\right)
\right].
\label{eq:reuse_risk}
\end{equation}
This population quantity is distinct from estimating a downstream rule from
finite target supervision.

For deterministic $h$, exact reuse under unrestricted deterministic decoding
means that there exist
$d_h:\mathcal Z\rightarrow\mathcal Y_t$
and
$\Omega_t\subseteq\operatorname{supp}(P_t)$
with $P_t(\Omega_t)=1$ such that
\begin{equation}
h(x)=d_h(U(x)),
\qquad
\forall x\in\Omega_t.
\label{eq:exact_reuse}
\end{equation}
Population-risk maps are assumed measurable. The exact existence results below
are deterministic statements; the corresponding measurability qualification
is given in the Appendix.

In our primary experiments, source learning uses pairwise relations. Let
$q_\psi^{\mathrm{rel}}:
\mathcal Z\times\mathcal Z\rightarrow\mathcal R$
be the relational source head, with population objective
\begin{equation}
\mathcal L_{\mathrm{rel}}(\theta,\psi)
=
\mathbb E_{(x_i,x_j,r_{ij})}
\left[
\ell_{\mathrm{rel}}
\left(
q_\psi^{\mathrm{rel}}
\left(
U_\theta(x_i),U_\theta(x_j)
\right),
r_{ij}
\right)
\right].
\label{eq:relation_loss}
\end{equation}
We write
$\widehat{\mathcal L}_{\mathrm{src}}$
for the empirical source-learning loss actually optimized in the corresponding
experimental setting.


\subsection{The Task-Relevant State of a Future-Task Family}
\label{sec:task_relative_state}

Learning representations that support multiple related tasks has a long
history in inductive-bias learning and multitask representation learning
\citep{1622248.1622254,2946645.3007034}.
We instead treat the future-task family as specified and ask for the coarsest
deterministic state sufficient for every task in that family.

Let
$\mathcal H=\{h_1,\ldots,h_M\}$
be a finite family of deterministic future tasks. Define
\begin{equation}
x\sim_{\mathcal H}x'
\quad\Longleftrightarrow\quad
h(x)=h(x')
\;\;
\forall h\in\mathcal H,
\label{eq:task_family_equivalence}
\end{equation}
and
\begin{equation}
s_{\mathcal H}^{\star}(x)
=
[x]_{\mathcal H},
\qquad
\mathcal S_{\mathcal H}^{\star}
=
\mathcal X/{\sim_{\mathcal H}}.
\label{eq:minimal_reusable_state}
\end{equation}

A deterministic state
$s:\mathcal X\rightarrow\mathcal S$
is sufficient for $\mathcal H$ if every
$h\in\mathcal H$
factors through $s$. By construction, every task factors through
$s_{\mathcal H}^{\star}$, and every sufficient $s$ satisfies
\begin{equation}
s_{\mathcal H}^{\star}
=
r\circ s
\label{eq:minimal_state_coarsening}
\end{equation}
for some deterministic
$r:\mathcal S\rightarrow\mathcal S_{\mathcal H}^{\star}$.
Thus, up to relabeling,
$s_{\mathcal H}^{\star}$
is the coarsest deterministic state sufficient for the family.
This construction is related to prior notions of sufficient and minimal
representations
\citep{8503149,NEURIPS2020_4c2e5eaa,9880437},
but here sufficiency is defined relative to an explicit family of future
tasks.

\paragraph{Lemma 1 (Task-Family Refinement).}
If
$\mathcal H_1\subseteq\mathcal H_2$,
then there exists a deterministic map $r_{21}$ such that
\begin{equation}
s_{\mathcal H_1}^{\star}
=
r_{21}\circ s_{\mathcal H_2}^{\star}.
\label{eq:task_family_refinement}
\end{equation}
Hence enlarging the future-task family cannot make its required state strictly
coarser.

\paragraph{Corollary 1 (Insufficient-State Impossibility).}
If
$s_{\mathcal H}^{\star}$
does not factor through a state
$s:\mathcal X\rightarrow\mathcal S$,
then there exist
$x_a,x_b\in\mathcal X$
and
$h\in\mathcal H$
such that
\begin{equation}
s(x_a)=s(x_b),
\qquad
h(x_a)\neq h(x_b).
\label{eq:insufficient_state_collision}
\end{equation}
No deterministic decoder acting only through $s$ can recover that task exactly
on a support containing both inputs. A strictly positive distributional error
lower bound additionally requires non-degenerate conditional target labels on
a set of positive state measure; see the Appendix.


\subsection{Exact Reuse Through an Effective Interface}
\label{sec:reuse_criterion}

The task-relevant state specifies which distinctions must remain available.
Exact reuse also depends on whether those distinctions admit a consistent
interpretation through the representation exposed to a downstream rule.
Throughout this subsection, decoding is unrestricted and deterministic.

Prior identifiability results study when learned representations are
recoverable or aligned only up to shared transformation classes
\citep{pmlr-v108-khemakhem20a,pmlr-v139-roeder21a,ICLR2025_5ee7ed60,ICLR2026_f67e5f99}.
Related work on representation comparison likewise makes explicit that the
notion of equivalence depends on which coordinate transformations are treated
as immaterial
\citep{pmlr-v97-kornblith19a}.
Given the representation interface exposed to reuse, we ask whether all
admissible realizations admit one common downstream interpretation.

For each $x\in\mathcal X$, let
$\mathcal G(x)\subseteq\operatorname{Bij}(\mathcal Z)$
be a nonempty family of admissible reversible coordinate transformations with
$\mathrm{id}\in\mathcal G(x)$. The admissible family is fixed by the reuse
setting independently of observed outcomes.
Define the \emph{effective interface}
\begin{equation}
\mathcal E_U^{\mathcal G}(x)
=
\{g(U(x)):g\in\mathcal G(x)\}.
\label{eq:effective_interface}
\end{equation}
A target $h$ is exactly reusable over this interface if, for some full-measure
$\Omega_t\subseteq\operatorname{supp}(P_t)$, one decoder $d$ satisfies
\begin{equation}
d(z)=h(x),
\qquad
\forall x\in\Omega_t,\;
\forall z\in\mathcal E_U^{\mathcal G}(x).
\label{eq:effective_exact_reuse}
\end{equation}

\paragraph{Theorem 1 (Effective-Interface Reuse Criterion).}
A deterministic target $h$ is exactly reusable over the effective interface
if and only if there exists a full-measure
$\Omega_t\subseteq\operatorname{supp}(P_t)$
such that
\begin{equation}
\mathcal E_U^{\mathcal G}(x_a)
\cap
\mathcal E_U^{\mathcal G}(x_b)
\neq\varnothing
\quad\Longrightarrow\quad
h(x_a)=h(x_b),
\qquad
\forall x_a,x_b\in\Omega_t.
\label{eq:reuse_criterion}
\end{equation}
Failure occurs exactly when two inputs requiring different target outputs can
present a common effective representation to the shared decoder. The proof and
measurability qualification are given in Appendix~\ref{app:theorem1_proof}.

\paragraph{Structural accessibility and coordinate consistency.}
Define
\begin{align}
\mathcal A_{U,h}(\Omega_t)
&=
\left\{
(x_a,x_b)\in\Omega_t^2:
U(x_a)=U(x_b),\;
h(x_a)\neq h(x_b)
\right\},
\label{eq:accessibility_conflict_set}\\
\mathcal K_{U,h}^{\mathcal G}(\Omega_t)
&=
\left\{
(x_a,x_b)\in\Omega_t^2:
U(x_a)\neq U(x_b),\;
\mathcal E_U^{\mathcal G}(x_a)
\cap
\mathcal E_U^{\mathcal G}(x_b)
\neq\varnothing,\;
h(x_a)\neq h(x_b)
\right\}.
\label{eq:coordinate_conflict_set}
\end{align}
Structural accessibility holds when
$\mathcal A_{U,h}(\Omega_t)=\varnothing$,
and coordinate consistency when
$\mathcal K_{U,h}^{\mathcal G}(\Omega_t)=\varnothing$.
Because
$\mathrm{id}\in\mathcal G(x)$,
these distinguish the two ways the criterion can fail: a required distinction
may already be collapsed by $U$, or distinct base representations may become
incompatible through their effective interfaces.

For the identity-interface special case
$\mathcal G(x)=\{\mathrm{id}\}$,
coordinate conflicts vanish and structural accessibility alone is sufficient
for unrestricted exact reuse. The graded conditional-accessibility quantity
used in Section~\ref{sec:experiments} is an empirical intervention diagnostic,
not this binary exact condition; no universal monotonicity with restricted
reuse follows from Theorem~1.

\paragraph{Corollary 2 (Family Reuse as Task-State Recovery).}
For finite $\mathcal H$ under a common target distribution, exact reuse of
every task over a common full-measure support
$\Omega_{\mathcal H}$
is equivalent to recovery of the family state:
\begin{equation}
\begin{aligned}
&\forall h\in\mathcal H,\ \exists d_h:
\ d_h(z)=h(x)
\\[-1mm]
&\qquad
\forall x\in\Omega_{\mathcal H},
\ \forall z\in\mathcal E_U^{\mathcal G}(x)
\\
&\Longleftrightarrow\
\exists\rho:
\ \rho(z)=s_{\mathcal H}^{\star}(x)
\quad
\forall x\in\Omega_{\mathcal H},
\ \forall z\in\mathcal E_U^{\mathcal G}(x).
\end{aligned}
\label{eq:family_effective_state_recovery}
\end{equation}

\paragraph{Proposition 1 (Shared-Coordinate Invariance).}
For any bijection
$\tau:\mathcal Z\rightarrow\mathcal Z$, let
\[
\widetilde U=\tau\circ U,
\qquad
\widetilde{\mathcal G}(x)
=
\{\tau\circ g\circ\tau^{-1}:g\in\mathcal G(x)\}.
\]
Then
\[
\mathcal E_{\widetilde U}^{\widetilde{\mathcal G}}(x)
=
\tau\!\left(
\mathcal E_U^{\mathcal G}(x)
\right),
\]
and
$d\circ\tau^{-1}$
decodes the transformed representation whenever $d$ decodes the original
one. Hence a globally shared bijective reparameterization preserves
unrestricted exact reuse.

For a restricted decoder class $\mathcal D$, define
\[
\widetilde{\mathcal D}
=
\{d\circ\tau^{-1}:d\in\mathcal D\}.
\]
Comparing
$(U,\mathcal D)$
with
$(\widetilde U,\widetilde{\mathcal D})$
preserves the corresponding reuse risk. If the same fixed decoder class is
used before and after reparameterization, closure under the corresponding
forward and inverse compositions is a sufficient condition for the same
invariance.

For
$x_{1:m}=(x_1,\ldots,x_m)$,
define
\[
U^{(m)}(x_{1:m})
=
\left(
U(x_1),\ldots,U(x_m)
\right).
\]
Let
$\mathcal G^{(m)}(x_{1:m})$
act componentwise:
\[
\mathcal G^{(m)}(x_{1:m})
=
\left\{
(g_1,\ldots,g_m):
g_j\in\mathcal G(x_j)
\right\},
\]
with
\[
(g_1,\ldots,g_m)
\left(
U(x_1),\ldots,U(x_m)
\right)
=
\left(
g_1(U(x_1)),\ldots,g_m(U(x_m))
\right).
\]
The joint effective interface is therefore
\begin{equation}
\mathcal E_{U^{(m)}}^{\mathcal G^{(m)}}(x_{1:m})
=
\prod_{j=1}^{m}
\mathcal E_U^{\mathcal G}(x_j).
\label{eq:joint_effective_interface}
\end{equation}


\subsection{Learning for Reuse Under Candidate-State and Decoder Constraints}
\label{sec:structure_preserving_learning}

Multitask learning commonly uses related auxiliary objectives to shape a
shared representation
\citep{Caruana_1997,2946645.3007034}.
Here the candidate state is tied explicitly to an anticipated future-task
family and is related to the corresponding task-relevant state through
Eq.~(\ref{eq:candidate_refines_minimal_state}).

Let
$s:\mathcal X\rightarrow\mathcal S$
be a candidate state. For an $m$-input downstream setting, let
\[
\Phi_m
\subseteq
\left\{
\phi:\mathcal S^m\rightarrow\mathcal Y_t
\right\}
\]
be a class of composition rules and define
\begin{equation}
\mathcal H_S^{(m)}
=
\left\{
h_\phi:
h_\phi(x_{1:m})
=
\phi\!\left(
s(x_1),\ldots,s(x_m)
\right),
\;
\phi\in\Phi_m
\right\}.
\label{eq:structure_induced_task_family}
\end{equation}
Writing
\[
s^{(m)}(x_{1:m})
=
\left(
s(x_1),\ldots,s(x_m)
\right),
\]
any finite anticipated family
$\mathcal H\subseteq\mathcal H_S^{(m)}$
satisfies
\begin{equation}
s_{\mathcal H}^{\star}
=
r\circ s^{(m)}
\label{eq:candidate_refines_minimal_state}
\end{equation}
for some
$r:\mathcal S^m\rightarrow\mathcal S_{\mathcal H}^{\star}$.
Thus the tuple extension of the candidate state refines the task-relevant
state required by the anticipated family.

Let
$\mathcal D_{\mathrm{str}}$
be a class of readouts
$\rho:\mathcal Z\rightarrow\mathcal S$.
Targeting candidate-state recoverability during source learning corresponds to
\begin{equation}
\mathcal R_{\mathrm{str}}
\left(
U;s,\mathcal D_{\mathrm{str}}
\right)
=
\inf_{\rho\in\mathcal D_{\mathrm{str}}}
\mathbb E_{X\sim P_s^X}
\left[
\ell_{\mathrm{str}}
\left(
\rho(U(X)),s(X)
\right)
\right].
\label{eq:structural_recoverability_risk}
\end{equation}
For the discrete states used experimentally, we optimize
\begin{equation}
\widehat{\mathcal R}_{\mathrm{str}}(\theta,\omega)
=
-\frac{1}{N}
\sum_{i=1}^{N}
\log
p_\omega
\left(
s_i\mid U_\theta(x_i)
\right)
\label{eq:empirical_structural_risk}
\end{equation}
and train with
\begin{equation}
\mathcal L_{\mathrm{SPR}}(\theta,\psi,\omega)
=
\widehat{\mathcal L}_{\mathrm{src}}(\theta,\psi)
+
\lambda_{\mathrm{reuse}}
\widehat{\mathcal R}_{\mathrm{str}}(\theta,\omega).
\label{eq:spr_training}
\end{equation}
SPR targets the structural-accessibility side of the theory. It does not
directly enforce coordinate consistency, and the structural readout is
discarded before reuse evaluation.

\paragraph{Proposition 2 (Candidate-State Recovery Suffices for Reuse).}
Let
$P_t^{(m)}$
be a downstream distribution on $\mathcal X^m$.
Suppose there exist
$\Omega_t\subseteq\mathcal X$
and
$\rho:\mathcal Z\rightarrow\mathcal S$
such that
\begin{equation}
P_t^{(m)}(\Omega_t^m)=1
\label{eq:target_product_support}
\end{equation}
and
\begin{equation}
\rho(z)=s(x),
\qquad
\forall x\in\Omega_t,\;
\forall z\in\mathcal E_U^{\mathcal G}(x).
\label{eq:effective_state_recovery}
\end{equation}
Then every
$h_\phi\in\mathcal H_S^{(m)}$
is exactly reusable over the joint effective interface via
\begin{equation}
d_\phi(z_1,\ldots,z_m)
=
\phi\!\left(
\rho(z_1),\ldots,\rho(z_m)
\right).
\label{eq:composed_reuse_decoder}
\end{equation}
The proposition assumes candidate-state recovery on the target effective
interface. The source-domain SPR objective promotes such recoverability during
training but does not itself establish this target-domain condition.

\paragraph{Restricted reuse under decoder constraints.}
For restricted reuse, we fix the realized coordinates of $U$ and separate
decoder-class restriction from coordinate ambiguity. Let
\[
\mathcal Q_s
\subseteq
\left\{
a:\mathcal Z\rightarrow\mathcal S
\right\}
\]
be a nonempty class of target-domain candidate-state readouts, and let
\[
\mathcal D_m
\subseteq
\left\{
d:\mathcal Z^m\rightarrow\mathcal Y_t
\right\}
\]
be a nonempty downstream decoder class.

For
$X_{1:m}\sim P_t^{(m)}$,
define under $0$--$1$ loss
\begin{align}
\epsilon_s(a)
&=
\frac{1}{m}
\sum_{j=1}^{m}
\Pr
\left[
a(U(X_j))
\neq
s(X_j)
\right],
\label{eq:state_recoverability_error}\\
\eta_\phi(a)
&=
\inf_{d\in\mathcal D_m}
\Pr
\left[
d(U^{(m)}(X_{1:m}))
\neq
\phi\!\left(
a(U(X_1)),\ldots,a(U(X_m))
\right)
\right],
\label{eq:restricted_composition_defect}\\
\alpha_\phi
&=
\inf_{d\in\mathcal D_m}
\Pr
\left[
d(U^{(m)}(X_{1:m}))
\neq
h_\phi(X_{1:m})
\right].
\label{eq:restricted_population_reuse_error}
\end{align}
For a fixed $a$, write
\[
G_{\phi,a}
=
\phi\!\left(
a(U(X_1)),\ldots,a(U(X_m))
\right).
\]

\paragraph{Proposition 3 (Restricted-Reuse Sandwich).}
For every
$a\in\mathcal Q_s$,
\begin{equation}
\left|
\alpha_\phi-\eta_\phi(a)
\right|
\le
\Pr
\left[
G_{\phi,a}
\neq
h_\phi(X_{1:m})
\right]
\le
m\,\epsilon_s(a).
\label{eq:restricted_reuse_pointwise_bound}
\end{equation}
Consequently,
\begin{equation}
\sup_{a\in\mathcal Q_s}
\left[
\eta_\phi(a)-m\epsilon_s(a)
\right]_+
\le
\alpha_\phi
\le
\inf_{a\in\mathcal Q_s}
\left[
\eta_\phi(a)+m\epsilon_s(a)
\right].
\label{eq:restricted_reuse_bound}
\end{equation}
The same readout couples recoverability and restricted composition, so better
candidate-state recovery need not monotonically improve restricted reuse;
Appendix~\ref{app:recoverability_counterexample} gives a counterexample.
For finite $\mathcal D_m$, fitting from $n$ i.i.d.\ target tuples adds the
standard
$O(\sqrt{\log|\mathcal D_m|/n})$
estimation term
(Appendix~\ref{app:finite_sample_reuse}).
SPR optimizes a source-domain surrogate and directly optimizes neither
$\eta_\phi$, this finite-sample term, nor coordinate consistency;
Section~\ref{sec:experiments} evaluates the corresponding empirical behavior.


\section{Experiments}
\label{sec:experiments}

The experiments follow the three questions raised by the theory. First, can
representations with matched source-task performance nevertheless support
substantially different future tasks after the source head is removed?
Second, when reuse differs, can the failure be separated into loss of
task-relevant distinctions and loss of a shared downstream interpretation?
Third, if the future-task family is anticipated, can source learning preserve
the required distinctions so that later tasks reuse the representation more
effectively under limited target supervision?

C0 (\emph{Source-only}) uses only the source objective. C1 (\emph{SPR}) adds
Eq.~(\ref{eq:spr_training}). C4 (\emph{Globally relabeled SPR}) uses the same
objective after one fixed bijective relabeling of candidate-state labels;
representation coordinates are unchanged. Unless stated otherwise, reuse
evaluation discards the source head, freezes the encoder, and fits a new
downstream rule. Implementation details, admissibility criteria, and statistical
procedures are given in the Appendix.


\subsection{Source-Task Success Does Not Determine Representation Reuse}
\label{sec:task_success_reuse}

This section tests whether source-task success can rank learned
representations for future reuse; it does not assume that SPR should improve
every replacement task family. Under a task-relative account, representations
with matched source performance can differ in reuse, and their ordering need
not be preserved across future-task families that require different
distinctions.

\begin{figure}[t]
    \centering
    \begin{minipage}[t]{0.47\linewidth}
        \centering
        \includegraphics[width=\linewidth]
        {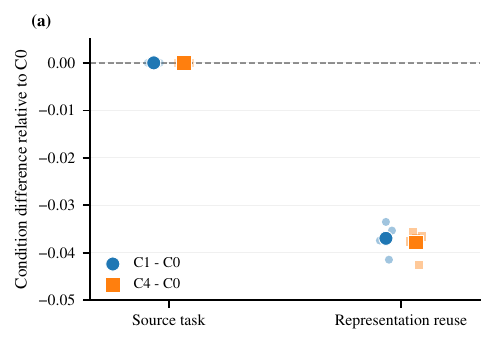}
    \end{minipage}
    \hfill
    \begin{minipage}[t]{0.47\linewidth}
        \centering
        \includegraphics[width=\linewidth]
        {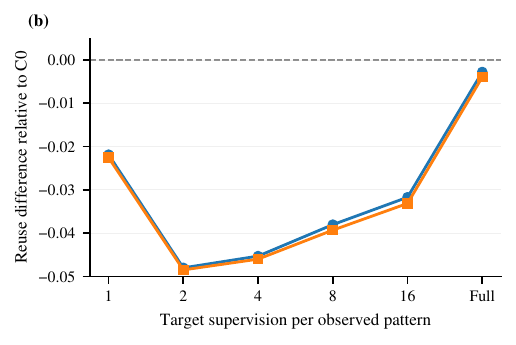}
    \end{minipage}
    \caption{
    \textbf{Source-task performance does not determine reuse.}
    \textbf{(a)} Paired reuse differences under exact source matching.
    \textbf{(b)} Mean reuse differences as target supervision increases;
    the dashed line denotes zero.
    }
    \label{fig:claim1_shapes3d}
\end{figure}

\paragraph{Exact source matching and target supervision.}
On Shapes3D \citep{pmlr-v80-kim18b}, five seeds for each of C0, C1, and C4
all achieve exactly $1.0000$ source validation accuracy. We use
\emph{low-shot reuse} for the predefined aggregate over the five
limited-supervision budgets $K\in\{1,2,4,8,16\}$.
Across $1004$ admissible novel-rule evaluations, mean low-shot reuse is
$0.6410$, $0.6040$, and $0.6031$, respectively; the paired C1--C0 and C4--C0 differences are
$-0.0370$ and $-0.0379$, with the same sign for all five seeds
(Figure~\ref{fig:claim1_shapes3d}(a)). With full target support, these
differences contract to $-0.0028$ and $-0.0040$
(Figure~\ref{fig:claim1_shapes3d}(b)). The source-matched representations differ in reuse, with a larger separation
under limited target supervision.

\paragraph{Extension across data and source-learning settings.}
On CIFAR-100 \citep{krizhevsky2009learning}, nine matched representation pairs
have a mean absolute source-performance gap of $0.0036$, while C1 exceeds C0
in low-shot reuse by $0.1718$. Under an ordinary supervised-classification
source task, source accuracy is similarly close ($0.7535$ versus $0.7602$),
while the held-out-rule reuse difference is $+0.0768$. Source-matched representations therefore need not have matched reuse, and
their ordering can change across future-task families that require different
distinctions. In our experiments, preserving a candidate state helps tasks
that depend on that state but does not make the resulting representation
uniformly better for unrelated replacement families. This family dependence is
consistent with the task-relative account.


\subsection{Structural Accessibility and Coordinate Consistency}
\label{sec:reuse_mechanism}

We test finite-supervision counterparts of the two exact-reuse requirements in
Section~\ref{sec:reuse_criterion}. These measurements are empirical diagnostics,
not the exact conflict sets defined by the theory.

\begin{figure}[t]
    \centering
    \includegraphics[width=0.90\linewidth]
    {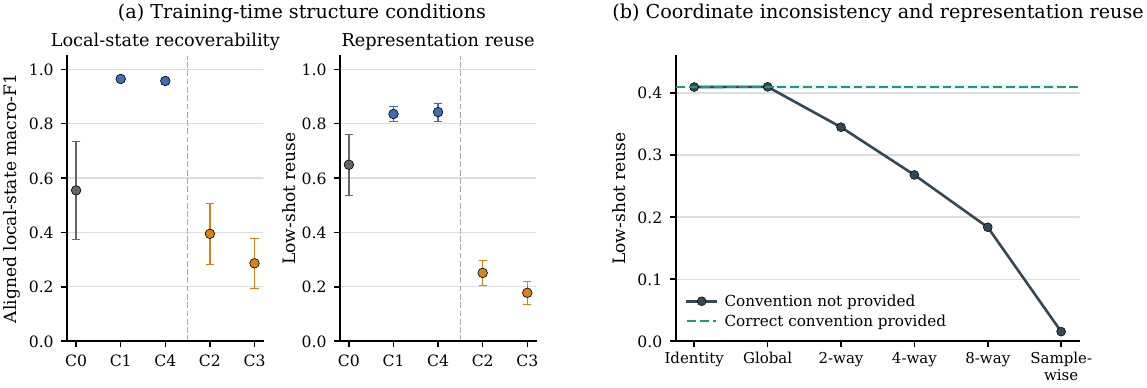}
    \caption{
    \textbf{Accessibility and coordinate tests.}
    \textbf{(a)} State recoverability and low-shot reuse for C0, C1, C4,
    Random auxiliary (C2), and Shuffled-state (C3).
    \textbf{(b)} Reuse under reversible coordinate interventions;
    increasingly heterogeneous sample-dependent conventions reduce blind
    reuse, while convention-aware inversion returns reuse to the Identity
    condition. Error bars show one standard deviation.
    }
    \label{fig:reuse_mechanisms}
\end{figure}

\paragraph{Candidate-state recoverability.}
Across five matched seeds, C1 raises aligned held-out state macro-F1 from
$0.5547\pm0.1807$ to $0.9646\pm0.0052$, a paired gain of $0.4099$
(95\% CI $[0.2719,\,0.5479]$). Low-shot reuse rises from
$0.6491\pm0.1130$ to $0.8356\pm0.0274$, with a paired gain of $0.1866$
($[0.1012,\,0.2875]$), positive in all five seeds
(Figure~\ref{fig:reuse_mechanisms}(a)). C4 reaches similar recoverability
($0.9572\pm0.0109$) and differs from C1 in reuse by $-0.0072$
($[-0.0375,\,0.0203]$), consistent with insensitivity to a global relabeling
of candidate-state symbols. Random-auxiliary and Shuffled-state controls have
lower recoverability and reuse; C1 exceeds them in reuse by $0.5845$ and
$0.6576$, respectively, with the same sign in all five seeds. The aligned
macro-F1 is diagnostic only and is never supplied to the downstream reuse
evaluator.

\paragraph{Recoverable information can still be unusable.}
We apply reversible coordinate transformations to the same frozen
representation. Reuse is $0.4096$ under Identity and $0.4099$ under one
globally shared reparameterization, but falls to $0.3449$, $0.2679$,
$0.1836$, and $0.0154$ under 2-way, 4-way, 8-way, and sample-wise
conventions, respectively (Figure~\ref{fig:reuse_mechanisms}(b)). The same
ordering holds for all three source-training seeds. Every transformation is
invertible. Supplying the applicable convention and applying its inverse
restores reuse to $0.4096$ in every condition, with zero reconstruction
difference from the common-reference representation. The information needed
for reuse therefore remains recoverable, yet a shared downstream rule fails
when different samples expose that information through incompatible coordinate
conventions.

\paragraph{Prospective natural-image replication.}
We test five protocol-defined CIFAR-100 conditions Q1--Q5 with normalized
conditional-accessibility values approximately $0$, $0.4182$, $0.6555$,
$0.8277$, and $1.0000$. These are predefined intervention levels, not post hoc
estimates of information retained by $U$; Q1$=0$ refers only to the controlled
conditional component. Construction and normalization details are given in the
Appendix.

Held-out reuse increases from Q1 through Q5 for both architectures. ResNet-18
rises from $0.2670$ to $0.3213$ ($\Delta_{5-1}=+0.0542$), and DenseNet-121
from $0.2811$ to $0.3261$ ($\Delta_{5-1}=+0.0450$). Full condition-wise values
are reported in Table~\ref{tab:app_conditional_accessibility} of
Appendix~\ref{app:conditional_accessibility_results}. At the representation-seed
level, all five accessibility--reuse slopes are positive for each architecture.
Mean slopes are $0.0516$ for ResNet-18
(95\% exact-bootstrap CI $[0.0485,\,0.0547]$) and $0.0460$ for DenseNet-121
($[0.0440,\,0.0477]$). The seed-level estimand and resampling procedure are
given in the Appendix.

The result is specific to the predefined accessibility hierarchy, future-task
family, and intervention construction. Proposition~3 does not imply a universal
monotonic relationship between accessibility and restricted reuse.


\subsection{Structure-Preserving Learning Improves Reuse under Limited Target Supervision}
\label{sec:structure_preserving_reuse}

\begin{figure}[t]
    \centering
    \includegraphics[width=0.90\linewidth]
    {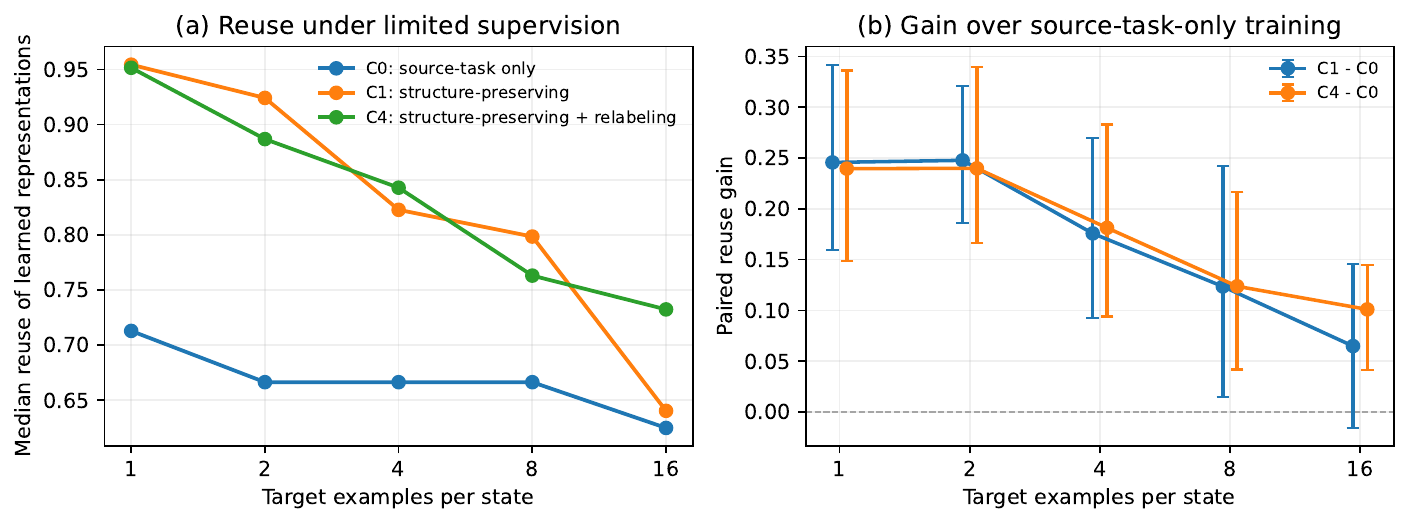}
    \caption{
    \textbf{Structure preservation improves reuse primarily under limited target
    supervision.}
    \textbf{(a)} Reuse across the predefined target-supervision conditions for
    C0, C1, and C4.
    \textbf{(b)} Paired seed-level gains over C0: C1 improves low-shot reuse by
    $0.1715$, whereas the full-support gain is $0.0241$ and its confidence
    interval includes zero. C4 shows the same supervision-localized pattern.
    Panel (a) is descriptive; panel (b) reports the paired estimand with bootstrap
    95\% confidence intervals.
    }
    \label{fig:structure_preserving_reuse}
\end{figure}

On held-out tasks whose task-relevant states are refined by the candidate
state promoted during source learning, the reuse gain is concentrated under
limited target supervision. Across
$K\in\{1,2,4,8,16\}$ examples per state, the paired C1--C0 low-shot gain is
$0.1715$ (95\% CI $[0.0986,\,0.2591]$), and the C4--C0 gain is
$0.1771$ ($[0.1172,\,0.2593]$); both are positive in all five matched seeds
(Figure~\ref{fig:structure_preserving_reuse}). Under full target support, the
corresponding gains are $0.0241$ ($[-0.0260,\,0.0992]$) and $0.0300$
($[-0.0201,\,0.0990]$).

These tasks instantiate $\mathcal H_S^{(m)}$ from
Eq.~(\ref{eq:structure_induced_task_family}): they share the candidate state
$s$ and vary the composition rule $\phi$, so their task-relevant states are
refined by $s^{(m)}$ as in
Eq.~(\ref{eq:candidate_refines_minimal_state}). The primary paired analysis
retains $21$ of $24$ registered family--rule evaluations after a
representation-independent admissibility screen; registry construction and
class support are given in the Appendix.

Let
$\Delta_{\mathrm{LS}}(C)$
and
$\Delta_{\mathrm{full}}(C)$
denote paired gains under the predefined limited-supervision and full-support
regimes, respectively. Define
\begin{equation}
\Gamma(C)
=
\Delta_{\mathrm{LS}}(C)
-
\Delta_{\mathrm{full}}(C).
\label{eq:supervision_localized_gain}
\end{equation}
We obtain
$\Gamma(\mathrm{C1})=0.1474$
(95\% CI $[0.1044,\,0.1905]$)
and
$\Gamma(\mathrm{C4})=0.1471$
($[0.1193,\,0.1749]$).
Both $\Gamma$ values are positive, while both full-support intervals include
zero. The advantage of C1 and C4 over C0 is concentrated under limited target
supervision. In this setting, distinctions already accessible from the frozen
representation matter most when the downstream learner has few labeled target
examples.

The direct C1--C4 low-shot contrast is $-0.0056$
(95\% CI $[-0.0284,\,0.0173]$), with no consistent direction across seeds,
consistent with insensitivity to a global relabeling of candidate-state
symbols. The gain reported here is specific to the predefined held-out tasks
that factor through $s$. As Section~\ref{sec:task_success_reuse} shows,
families requiring different distinctions can induce a different ordering of
the same source-matched representations.


\section{Conclusion}
\label{sec:conclusion}

A representation can solve its source task well and still be difficult to
reuse once the source head is removed. Our framework separates two failures:
distinctions required by later tasks may no longer be recoverable, or they may
remain recoverable but be exposed through coordinates that a shared downstream
rule cannot interpret consistently. The reversible coordinate experiment
isolates the second case: reuse falls under sample-dependent conventions and
returns when the applicable convention is inverted.

The practical consequence is clearest when target supervision is limited. The
predefined accessibility ordering repeats with both ResNet-18 and DenseNet-121,
and preserving a candidate state that refines the held-out task requirements
improves low-shot reuse while the full-support intervals include zero. The
benefit is family-specific: future tasks requiring different distinctions can
change the ordering of source-matched representations. Representation reuse
should therefore be evaluated relative to the tasks that follow source
learning---whether the distinctions they require remain available in a form
that a shared downstream rule can use.

\paragraph{AI use statement.}
Generative AI tools were used to assist with literature organization, translation, language editing, and limited implementation support during the preparation of this work. All scientific claims, theoretical statements, experimental designs, implementations, numerical results, and final manuscript content were reviewed and verified by the authors. The authors take full responsibility for the content of this paper.

\clearpage

\bibliography{references}
\bibliographystyle{plainnat}


\appendix

\section{Formal Details and Proofs}
\label{app:formal_details}


\subsection{Minimal State for a Finite Future-Task Family}
\label{app:minimal_state}

Let $\mathcal H$ be a finite family of deterministic target tasks on
$\mathcal X$. Define
\[
x\sim_{\mathcal H}x'
\quad\Longleftrightarrow\quad
h(x)=h(x')
\quad
\forall h\in\mathcal H.
\]
The corresponding task-relevant state is
\[
s_{\mathcal H}^{\star}(x)
=
[x]_{\mathcal H}.
\]

For every $h\in\mathcal H$, $h(x)$ is constant on each equivalence class.
Hence there exists a function $\phi_h$ such that
\[
h
=
\phi_h\circ s_{\mathcal H}^{\star}.
\]
Thus $s_{\mathcal H}^{\star}$ is sufficient for every task in
$\mathcal H$.

It is also the coarsest deterministic state with this property. Suppose
\[
s:\mathcal X\rightarrow\mathcal S
\]
supports every task in $\mathcal H$, so that for each $h\in\mathcal H$
there exists $\widetilde\phi_h$ satisfying
\[
h
=
\widetilde\phi_h\circ s.
\]
If
\[
s(x)=s(x'),
\]
then
\[
h(x)
=
\widetilde\phi_h(s(x))
=
\widetilde\phi_h(s(x'))
=
h(x')
\qquad
\forall h\in\mathcal H.
\]
Therefore
\[
x\sim_{\mathcal H}x',
\]
so $s_{\mathcal H}^{\star}$ is constant on every fiber of $s$.
Consequently, there exists a deterministic map $r$ such that
\[
s_{\mathcal H}^{\star}
=
r\circ s.
\]

Hence every deterministic state sufficient for the entire family refines
$s_{\mathcal H}^{\star}$. Any other coarsest deterministic sufficient state
induces the same partition of $\mathcal X$ and differs from
$s_{\mathcal H}^{\star}$ only by a bijective relabeling of its state values.

\paragraph{Task-family refinement.}
Let
\[
\mathcal H_1\subseteq\mathcal H_2.
\]
If
\[
s_{\mathcal H_2}^{\star}(x)
=
s_{\mathcal H_2}^{\star}(x'),
\]
then every task in $\mathcal H_2$ takes the same value on $x$ and $x'$.
Because
$\mathcal H_1\subseteq\mathcal H_2$,
the same is true for every task in $\mathcal H_1$, and therefore
\[
s_{\mathcal H_1}^{\star}(x)
=
s_{\mathcal H_1}^{\star}(x').
\]
Thus
$s_{\mathcal H_1}^{\star}$
is constant on every fiber of
$s_{\mathcal H_2}^{\star}$,
so there exists a deterministic map $r_{21}$ satisfying
\[
s_{\mathcal H_1}^{\star}
=
r_{21}\circ s_{\mathcal H_2}^{\star}.
\]
Enlarging the future-task family therefore cannot make its required state
strictly coarser.

\paragraph{Insufficient-state impossibility.}
Let
\[
s:\mathcal X\rightarrow\mathcal S.
\]
Suppose
$s_{\mathcal H}^{\star}$
does not factor through $s$.
Then
$s_{\mathcal H}^{\star}$
is not constant on at least one fiber of $s$, so there exist
$x_a,x_b\in\mathcal X$ such that
\[
s(x_a)=s(x_b),
\qquad
s_{\mathcal H}^{\star}(x_a)
\neq
s_{\mathcal H}^{\star}(x_b).
\]
By definition of
$s_{\mathcal H}^{\star}$,
there exists some
$h\in\mathcal H$
for which
\[
h(x_a)\neq h(x_b).
\]
Any deterministic decoder that observes only $s(x)$ must assign the same
output to $x_a$ and $x_b$, and therefore cannot recover that task exactly on
a target support containing both inputs.

This conclusion is pointwise. A strictly positive population-risk lower bound
requires the conflicting target labels to remain indistinguishable on a set
with positive probability under the target distribution.

\paragraph{Distributional Bayes obstruction.}
Let
\[
S=s(X),
\qquad
Y=h(X),
\]
where $\mathcal Y_t$ is finite and the relevant regular conditional
distribution exists. Define
\begin{equation}
B_s(h)
=
\mathbb E_S
\left[
1-
\max_{y\in\mathcal Y_t}
\Pr(Y=y\mid S)
\right].
\label{eq:app_bayes_obstruction}
\end{equation}
This is the Bayes error of predicting $h(X)$ from $s(X)$.
For every measurable decoder
$\delta:\mathcal S\rightarrow\mathcal Y_t$,
\[
\Pr[\delta(S)\neq Y]
\ge
B_s(h).
\]

Moreover, if
\[
U=f\circ s
\]
for a deterministic measurable map $f$, every measurable decoder acting
through $U$ is also a measurable function of $S$. Hence
\[
\inf_d
\Pr[d(U(X))\neq h(X)]
\ge
B_s(h).
\]

The quantity $B_s(h)$ is strictly positive precisely when
\[
1-\max_y\Pr(Y=y\mid S)>0
\]
on a set of state values having positive $P_S$-measure.
Thus positive distributional error requires non-degenerate conditional target
labels on a set of positive measure; it does not require any individual state
value to have positive probability.


\subsection{Proof of Theorem 1}
\label{app:theorem1_proof}

Let
$\Omega\subseteq\operatorname{supp}(P_t)$
satisfy
\[
P_t(\Omega)=1.
\]
For every $x\in\Omega$, let
\[
\mathcal E_U^{\mathcal G}(x)
=
\{g(U(x)):g\in\mathcal G(x)\}
\]
denote the effective interface.

We prove that exact reuse over the effective interface is possible if and
only if
\begin{equation}
\mathcal E_U^{\mathcal G}(x_a)
\cap
\mathcal E_U^{\mathcal G}(x_b)
\neq\varnothing
\quad\Longrightarrow\quad
h(x_a)=h(x_b)
\label{eq:app_effective_interface_condition}
\end{equation}
for all
$x_a,x_b\in\Omega$.

\paragraph{Necessity.}
Suppose there exists one deterministic decoder $d$ satisfying
\[
d(z)=h(x),
\qquad
\forall x\in\Omega,\quad
\forall z\in\mathcal E_U^{\mathcal G}(x).
\]
Take any $x_a,x_b\in\Omega$ such that
\[
z
\in
\mathcal E_U^{\mathcal G}(x_a)
\cap
\mathcal E_U^{\mathcal G}(x_b).
\]
Then
\[
h(x_a)
=
d(z)
=
h(x_b),
\]
which establishes
Eq.~(\ref{eq:app_effective_interface_condition}).

\paragraph{Sufficiency.}
Assume
Eq.~(\ref{eq:app_effective_interface_condition}).
Define
\[
\mathcal Z_{\Omega}
=
\bigcup_{x\in\Omega}
\mathcal E_U^{\mathcal G}(x).
\]
For
$z\in\mathcal Z_{\Omega}$,
choose any $x\in\Omega$ satisfying
\[
z\in\mathcal E_U^{\mathcal G}(x)
\]
and define
\[
d(z)=h(x).
\]
This definition is well posed. If the same $z$ belongs to the effective
interfaces of both $x_a$ and $x_b$, then
Eq.~(\ref{eq:app_effective_interface_condition}) gives
\[
h(x_a)=h(x_b).
\]
Extend $d$ arbitrarily from $\mathcal Z_\Omega$ to the remainder of
$\mathcal Z$. The resulting decoder satisfies
\[
d(z)=h(x),
\qquad
\forall x\in\Omega,\quad
\forall z\in\mathcal E_U^{\mathcal G}(x).
\]
Therefore exact reuse holds on a full-measure support.

\paragraph{Measurability qualification.}
The argument above establishes existence of an unrestricted deterministic
decoder as a set-theoretic exact-decoding statement.
The overlap condition alone does not, under arbitrary measurable-space
structures, guarantee that the induced decoder or an arbitrary extension is
measurable.

Accordingly, Theorem~1 is interpreted first as a deterministic exact-decoding
criterion on the relevant full-measure support. Whenever the result is
translated into a zero population-risk statement involving an expectation,
the induced decoder is additionally required to be measurable.
This regularity condition is separate from restriction to a particular
parametric decoder class.


\subsection{Accessibility and Coordinate Consistency}
\label{app:accessibility_consistency}

Recall
\[
\mathcal A_{U,h}(\Omega)
=
\left\{
(x_a,x_b)\in\Omega^2:
h(x_a)\neq h(x_b),\;
U(x_a)=U(x_b)
\right\},
\]
and
\[
\mathcal K_{U,h}^{\mathcal G}(\Omega)
=
\left\{
(x_a,x_b)\in\Omega^2:
\begin{array}{l}
h(x_a)\neq h(x_b),\\
U(x_a)\neq U(x_b),\\
\mathcal E_U^{\mathcal G}(x_a)
\cap
\mathcal E_U^{\mathcal G}(x_b)
\neq\varnothing
\end{array}
\right\}.
\]

Consider any target-inconsistent pair whose effective interfaces overlap.
If
\[
U(x_a)=U(x_b),
\]
the pair belongs to
$\mathcal A_{U,h}(\Omega)$.
Otherwise it belongs to
$\mathcal K_{U,h}^{\mathcal G}(\Omega)$.

Conversely, because
$\mathrm{id}\in\mathcal G(x)$,
\[
U(x_a)=U(x_b)
\quad\Longrightarrow\quad
\mathcal E_U^{\mathcal G}(x_a)
\cap
\mathcal E_U^{\mathcal G}(x_b)
\neq\varnothing.
\]
Hence
\[
\left\{
(x_a,x_b)\in\Omega^2:
\begin{array}{l}
h(x_a)\neq h(x_b),\\
\mathcal E_U^{\mathcal G}(x_a)
\cap
\mathcal E_U^{\mathcal G}(x_b)
\neq\varnothing
\end{array}
\right\}
=
\mathcal A_{U,h}(\Omega)
\;\dot\cup\;
\mathcal K_{U,h}^{\mathcal G}(\Omega).
\]

Theorem~1 is therefore equivalent to requiring both conflict sets to be empty
on the relevant full-measure support.
Structural accessibility asks whether a target-relevant distinction has
already been collapsed by the representation.
Coordinate consistency asks whether distinct retained representations remain
jointly interpretable by one shared decoder over the admissible interface.


\subsection{Task-Family Reuse and Recovery of the Task-Relevant State}
\label{app:task_family_reuse}

Let $\mathcal H$ be finite and suppose all tasks are evaluated under the same
target distribution $P_t$.

If every task
$h\in\mathcal H$
admits a full-measure support
$\Omega_h$
on which exact reuse holds, then
\[
\Omega_{\mathcal H}
=
\bigcap_{h\in\mathcal H}\Omega_h
\]
also satisfies
\[
P_t(\Omega_{\mathcal H})=1.
\]

First consider the identity interface.
If there exists
\[
\rho:
\mathcal Z
\rightarrow
\operatorname{range}(s_{\mathcal H}^{\star})
\]
such that
\[
\rho(U(x))
=
s_{\mathcal H}^{\star}(x)
\]
on $\Omega_{\mathcal H}$, then for each
$h\in\mathcal H$,
\[
h
=
\phi_h\circ s_{\mathcal H}^{\star},
\]
and therefore
\[
d_h
=
\phi_h\circ\rho
\]
exactly reuses the representation.

Conversely, suppose every task in $\mathcal H$ is exactly reusable.
If
\[
U(x)=U(x'),
\]
then every task-specific decoder must return the same value on $x$ and $x'$.
Thus
\[
h(x)=h(x')
\qquad
\forall h\in\mathcal H,
\]
so
\[
x\sim_{\mathcal H}x'.
\]
Hence
$s_{\mathcal H}^{\star}(x)$
is determined by $U(x)$, and a shared state readout exists on the common
support.

The same argument applies to the effective interface.
Exact reuse of every task in the finite family is equivalent to the existence
of one state readout satisfying
\[
\rho(z)
=
s_{\mathcal H}^{\star}(x),
\qquad
\forall x\in\Omega_{\mathcal H},
\quad
\forall z\in\mathcal E_U^{\mathcal G}(x).
\]

The quantifier structure is
\[
\forall h\in\mathcal H,\ \exists d_h,
\]
rather than the existence of one decoder implementing all distinct tasks
simultaneously.

When these deterministic constructions are interpreted through population
risks, the same measurability convention as in
Appendix~\ref{app:theorem1_proof}
applies.


\subsection{Proof of Proposition 1}
\label{app:prop1_proof}

Let
\[
\tau:\mathcal Z\rightarrow\mathcal Z
\]
be a globally shared bijection and define
\[
\widetilde U
=
\tau\circ U.
\]
For each $x$, define
\[
\widetilde{\mathcal G}(x)
=
\left\{
\tau\circ g\circ\tau^{-1}
:
g\in\mathcal G(x)
\right\}.
\]
Then
\begin{align*}
\mathcal E_{\widetilde U}^{\widetilde{\mathcal G}}(x)
&=
\left\{
(\tau\circ g\circ\tau^{-1})
\bigl(\tau(U(x))\bigr)
:
g\in\mathcal G(x)
\right\}
\\
&=
\left\{
\tau(g(U(x)))
:
g\in\mathcal G(x)
\right\}
\\
&=
\tau
\left(
\mathcal E_U^{\mathcal G}(x)
\right).
\end{align*}

Suppose $d$ exactly decodes the original effective interface.
Define
\[
\widetilde d
=
d\circ\tau^{-1}.
\]
For any
\[
\widetilde z
\in
\mathcal E_{\widetilde U}^{\widetilde{\mathcal G}}(x),
\]
there exists
$z\in\mathcal E_U^{\mathcal G}(x)$
with
$\widetilde z=\tau(z)$.
Thus
\[
\widetilde d(\widetilde z)
=
d(z)
=
h(x).
\]
The converse follows by applying the same argument with $\tau^{-1}$.
Therefore a globally shared bijective reparameterization preserves exact reuse
when the admissible ambiguity family is transformed consistently.

\paragraph{Restricted decoder classes.}
For a restricted decoder class $\mathcal D$, define
\[
\widetilde{\mathcal D}
=
\left\{
d\circ\tau^{-1}:d\in\mathcal D
\right\}.
\]
For the same target distribution and loss,
\[
\mathcal R_{\mathrm{reuse}}
\left(
\widetilde U;
h,
\widetilde{\mathcal D}
\right)
=
\mathcal R_{\mathrm{reuse}}
\left(
U;
h,
\mathcal D
\right).
\]
Thus restricted reuse is invariant when the decoder class is transported
together with the coordinate change.

If the same fixed decoder class is used before and after reparameterization,
closure under the corresponding forward and inverse compositions is a
sufficient condition for the same invariance.
It is not asserted to be necessary for a particular
representation--task pair.

When these identities are interpreted as population-risk statements,
$\tau$, $\tau^{-1}$, and the corresponding transported decoders are
additionally required to be measurable so that the induced random variables
and expectations are well defined. This is the same population-risk
regularity convention used in
Appendix~\ref{app:theorem1_proof}.


\subsection{Joint Effective Interfaces}
\label{app:joint_effective_interfaces}

For an $m$-input downstream task, let
\[
x_{1:m}
=
(x_1,\ldots,x_m)
\in\mathcal X^m
\]
and let
$P_t^{(m)}$
be the target distribution on $\mathcal X^m$.
No product or independence assumption is required.

Define
\[
U^{(m)}(x_{1:m})
=
(U(x_1),\ldots,U(x_m)).
\]
Let
$\mathcal G^{(m)}(x_{1:m})$
contain componentwise transformations
\[
g_{1:m}(z_1,\ldots,z_m)
=
(g_1(z_1),\ldots,g_m(z_m)),
\qquad
g_j\in\mathcal G(x_j).
\]
Then
\[
\mathcal E_{U^{(m)}}^{\mathcal G^{(m)}}(x_{1:m})
=
\left\{
g_{1:m}
\bigl(
U(x_1),\ldots,U(x_m)
\bigr)
:
g_{1:m}\in\mathcal G^{(m)}(x_{1:m})
\right\}.
\]
Because the transformation family acts componentwise,
\[
\mathcal E_{U^{(m)}}^{\mathcal G^{(m)}}(x_{1:m})
=
\prod_{j=1}^{m}
\mathcal E_U^{\mathcal G}(x_j).
\]

For a candidate state
\[
s:\mathcal X\rightarrow\mathcal S,
\]
the corresponding tuple state is
\[
s^{(m)}(x_{1:m})
=
(s(x_1),\ldots,s(x_m)).
\]
For
$\Phi_m\subseteq
\{\phi:\mathcal S^m\rightarrow\mathcal Y_t\}$,
the induced family is
\[
\mathcal H_S^{(m)}
=
\left\{
h_\phi:
h_\phi(x_{1:m})
=
\phi(s^{(m)}(x_{1:m})),
\quad
\phi\in\Phi_m
\right\}.
\]


\subsection{Proof of Proposition 2}
\label{app:prop2_proof}

Suppose there exist
$\Omega_t\subseteq\mathcal X$
with
\[
P_t^{(m)}(\Omega_t^m)=1
\]
and a readout
\[
\rho:\mathcal Z\rightarrow\mathcal S
\]
such that
\[
\rho(z)=s(x),
\qquad
\forall x\in\Omega_t,
\quad
\forall z\in\mathcal E_U^{\mathcal G}(x).
\]

Take any
$h_\phi\in\mathcal H_S^{(m)}$
and define
\[
d_\phi(z_1,\ldots,z_m)
=
\phi(
\rho(z_1),\ldots,\rho(z_m)
).
\]
For any
\[
x_{1:m}\in\Omega_t^m
\]
and
\[
z_{1:m}
\in
\mathcal E_{U^{(m)}}^{\mathcal G^{(m)}}(x_{1:m}),
\]
the componentwise product identity gives
\[
z_j
\in
\mathcal E_U^{\mathcal G}(x_j),
\qquad
j=1,\ldots,m.
\]
Hence
\[
\rho(z_j)=s(x_j)
\]
for every component and therefore
\[
d_\phi(z_{1:m})
=
\phi(
s(x_1),\ldots,s(x_m)
)
=
h_\phi(x_{1:m}).
\]
Thus every task in
$\mathcal H_S^{(m)}$
is exactly reusable over the joint effective interface.

The proposition assumes candidate-state recovery on the target effective
interface.
The SPR objective in
Eq.~(\ref{eq:spr_training})
promotes recoverability during source learning but does not itself imply this
target-domain condition.

When the deterministic decoder above is interpreted through population risk,
the measurability convention in
Appendix~\ref{app:theorem1_proof}
applies.


\subsection{Proof of Proposition 3: Restricted-Reuse Sandwich}
\label{app:restricted_reuse_bound}

Let
\[
X_{1:m}
=
(X_1,\ldots,X_m)
\sim
P_t^{(m)}
\]
and fix
$a\in\mathcal Q_s$.
Write
\[
Y
=
h_\phi(X_{1:m})
\]
and
\[
G
=
G_{\phi,a}
=
\phi
\left(
a(U(X_1)),
\ldots,
a(U(X_m))
\right).
\]

For every fixed
$d\in\mathcal D_m$,
\[
\left|
\mathbf 1\{d(U^{(m)}(X_{1:m}))\neq Y\}
-
\mathbf 1\{d(U^{(m)}(X_{1:m}))\neq G\}
\right|
\le
\mathbf 1\{Y\neq G\}.
\]
Taking expectations gives
\begin{equation}
|R_Y(d)-R_G(d)|
\le
\Pr[Y\neq G].
\label{eq:app_fixed_decoder_target_comparison}
\end{equation}

By definition,
\[
\alpha_\phi
=
\inf_{d\in\mathcal D_m}R_Y(d),
\qquad
\eta_\phi(a)
=
\inf_{d\in\mathcal D_m}R_G(d).
\]
A uniform inequality
$|R_Y(d)-R_G(d)|\le c$
for every $d$ implies
\[
\left|
\inf_dR_Y(d)-\inf_dR_G(d)
\right|
\le c.
\]
Therefore
\begin{equation}
|\alpha_\phi-\eta_\phi(a)|
\le
\Pr[
G_{\phi,a}
\neq
h_\phi(X_{1:m})
].
\label{eq:app_restricted_risk_comparison}
\end{equation}

Define
\[
\widehat S_j
=
a(U(X_j)),
\qquad
S_j
=
s(X_j).
\]
If all
$\widehat S_j=S_j$,
then
\[
G_{\phi,a}
=
h_\phi(X_{1:m}).
\]
Thus
\[
\{
G_{\phi,a}
\neq
h_\phi(X_{1:m})
\}
\subseteq
\bigcup_{j=1}^{m}
\{
\widehat S_j\neq S_j
\}.
\]
By the union bound,
\begin{align}
\Pr[
G_{\phi,a}
\neq
h_\phi(X_{1:m})
]
&\le
\sum_{j=1}^{m}
\Pr[
a(U(X_j))
\neq
s(X_j)
]
\\
&=
m\,\epsilon_s(a).
\label{eq:app_proxy_union_bound}
\end{align}
No independence among the components of $X_{1:m}$ is required.

Combining
Eqs.~(\ref{eq:app_restricted_risk_comparison})
and~(\ref{eq:app_proxy_union_bound})
gives
\[
\eta_\phi(a)-m\epsilon_s(a)
\le
\alpha_\phi
\le
\eta_\phi(a)+m\epsilon_s(a).
\]
Since
$\alpha_\phi\ge0$,
\[
[\eta_\phi(a)-m\epsilon_s(a)]_+
\le
\alpha_\phi.
\]
Taking the supremum of the lower bounds and the infimum of the upper bounds
over
$a\in\mathcal Q_s$
gives
\[
\sup_{a\in\mathcal Q_s}
[\eta_\phi(a)-m\epsilon_s(a)]_+
\le
\alpha_\phi
\le
\inf_{a\in\mathcal Q_s}
[\eta_\phi(a)+m\epsilon_s(a)].
\]

The same readout $a$ determines both terms.
Candidate-state recoverability and restricted composition therefore cannot in
general be optimized independently, and the result does not imply a monotone
ordering of representations by state-recovery error alone.


\subsection{Finite-Sample Extension for a Finite Decoder Class}
\label{app:finite_sample_reuse}

For
$d\in\mathcal D_m$,
define
\[
\mathcal R_\phi(d)
=
\Pr[
d(U^{(m)}(X_{1:m}))
\neq
h_\phi(X_{1:m})
].
\]
Then
\[
\alpha_\phi
=
\inf_{d\in\mathcal D_m}
\mathcal R_\phi(d).
\]

Let
\[
X_{1:m}^{(1)},\ldots,X_{1:m}^{(n)}
\overset{\mathrm{i.i.d.}}{\sim}
P_t^{(m)}
\]
and define
\[
\widehat{\mathcal R}_{\phi,n}(d)
=
\frac{1}{n}
\sum_{i=1}^{n}
\mathbf 1
\left\{
d(U^{(m)}(X_{1:m}^{(i)}))
\neq
h_\phi(X_{1:m}^{(i)})
\right\}.
\]
Let
\[
\widehat d_n
\in
\arg\min_{d\in\mathcal D_m}
\widehat{\mathcal R}_{\phi,n}(d).
\]

If $\mathcal D_m$ is finite, Hoeffding's inequality and a union bound imply
that with probability at least $1-\delta$,
\begin{equation}
\sup_{d\in\mathcal D_m}
\left|
\mathcal R_\phi(d)
-
\widehat{\mathcal R}_{\phi,n}(d)
\right|
\le
\Delta_n,
\qquad
\Delta_n
=
\sqrt{
\frac{
\log(2|\mathcal D_m|/\delta)
}{
2n
}
}.
\label{eq:app_uniform_convergence}
\end{equation}
On this event,
\[
\mathcal R_\phi(\widehat d_n)
\le
\alpha_\phi+2\Delta_n.
\]
Using the upper side of Proposition~3 gives
\begin{equation}
\mathcal R_\phi(\widehat d_n)
\le
\inf_{a\in\mathcal Q_s}
\left[
\eta_\phi(a)
+
m\epsilon_s(a)
\right]
+
\sqrt{
\frac{
2\log(2|\mathcal D_m|/\delta)
}{
n
}
}.
\label{eq:app_finite_sample_reuse_bound}
\end{equation}

The three terms have distinct roles:
candidate-state recovery, restricted composition, and finite-sample
estimation.
The bound does not imply that the difference between two learned
representations must increase as $n$ decreases.

The experimental target-supervision protocols use held-out patterns and do not
directly instantiate the i.i.d.\ quantity $n$ above.
Eq.~(\ref{eq:app_finite_sample_reuse_bound}) should therefore be read as a
conceptual finite-sample decomposition; it does not predict the
supervision-localized effect in
Section~\ref{sec:structure_preserving_reuse}.


\subsection{Better Candidate-State Recovery Need Not Improve Restricted Reuse}
\label{app:recoverability_counterexample}

Let
\[
S=(S_1,S_2)
\sim
\operatorname{Unif}(\{0,1\}^2)
\]
and let the candidate state be
\[
s(S)=S.
\]
Consider the target
\[
h(S)
=
S_1\oplus S_2
\]
and two representations
\[
U_0(S)
=
(S_1\oplus S_2,0),
\qquad
U_1(S)
=
(S_1,S_2).
\]

Define the optimal full-state recovery error
\[
\epsilon_s^\star(U)
=
\inf_a
\Pr[
a(U(S))\neq s(S)
],
\]
where the infimum ranges over deterministic state readouts.

The full state is exactly recoverable from $U_1$, so
\[
\epsilon_s^\star(U_1)=0.
\]
From $U_0$, only parity is available.
Conditional on either parity value, two full candidate states are equally
possible, so
\[
\epsilon_s^\star(U_0)=\frac12.
\]

Now restrict downstream decoders to affine threshold functions
\[
\mathcal D_{\mathrm{aff}}
=
\left\{
z\mapsto
\mathbf 1\{w^\top z+b\ge0\}
:
w\in\mathbb R^2,\;
b\in\mathbb R
\right\}.
\]
For $U_0$, the target is the first representation coordinate, so
\[
\mathcal R_{\mathrm{reuse}}
(
U_0;
h,
\mathcal D_{\mathrm{aff}}
)
=
0.
\]
For $U_1$, XOR on the Boolean square is not linearly separable.
The minimum affine-threshold error is
\[
\mathcal R_{\mathrm{reuse}}
(
U_1;
h,
\mathcal D_{\mathrm{aff}}
)
=
\frac14.
\]
Consequently,
\[
\epsilon_s^\star(U_1)
<
\epsilon_s^\star(U_0),
\]
while
\[
\mathcal R_{\mathrm{reuse}}
(
U_1;
h,
\mathcal D_{\mathrm{aff}}
)
>
\mathcal R_{\mathrm{reuse}}
(
U_0;
h,
\mathcal D_{\mathrm{aff}}
).
\]

For the singleton family
$\mathcal H=\{h\}$,
the minimal task-relevant state is, up to relabeling, parity itself.
Thus $U_0$ preserves exactly the distinction needed for the task, whereas
$U_1$ preserves a larger candidate state whose required composition is not
realizable by the restricted decoder class.

The counterexample concerns the interaction between candidate-state recovery
and decoder-class restriction. Here $U_1$ recovers the full candidate state
exactly, yet the required XOR composition is not realizable by
$\mathcal D_{\mathrm{aff}}$. Candidate-state recovery alone therefore does not
order representations by restricted reuse.


\section{Experimental Setting and Evaluation Protocols}
\label{app:experimental_protocols}

This section specifies the experimental conditions, frozen-representation
evaluation protocols, admissibility rules, diagnostic measurements, and
statistical estimands used in
Section~\ref{sec:experiments}.


\subsection{Experimental Conditions}
\label{app:conditions}

The experiments use five source-learning conditions.

\begin{table}[t]
\centering
\small
\setlength{\tabcolsep}{5pt}
\caption{
Source-learning conditions.
C2 and C3 are structural controls.
}
\label{tab:condition_definitions}
\begin{tabular}{clp{0.58\linewidth}}
\hline
ID & Name & Definition \\
\hline
C0 & Source-only &
Optimizes the source-task objective without structural supervision. \\

C1 & SPR &
Adds the structure-preserving objective in
Eq.~(\ref{eq:spr_training})
using candidate-state labels. \\

C2 & Random auxiliary &
Uses random three-state auxiliary targets in place of the candidate state. \\

C3 & Shuffled-state &
Permutes candidate-state targets across observations, breaking their
observation--state correspondence. \\

C4 & Globally relabeled SPR &
Uses the same structural supervision as C1 after one fixed, globally shared
bijective relabeling of candidate-state labels. \\
\hline
\end{tabular}
\end{table}

C4 changes only the symbolic state naming,
\[
s(x)
\longmapsto
\pi(s(x)),
\qquad
\pi:\mathcal S\rightarrow\mathcal S
\]
with $\pi$ bijective.
It does not transform representation coordinates and is therefore distinct
from Proposition~1 and from the coordinate intervention below.


\subsection{Candidate State and Held-Out Task Construction}
\label{app:heldout_task_construction}

In the controlled relational setting, the candidate state has three values,
\[
\mathcal S
=
\{
\mathrm{MATCH},
\mathrm{MISMATCH},
\mathrm{UNKNOWN}
\},
\]
abbreviated as
\[
M,\qquad D,\qquad U.
\]

For two inputs,
\[
s^{(2)}(x_1,x_2)
=
\left(
s(x_1),s(x_2)
\right)
\in
\mathcal S^2.
\]
Ordered configurations that differ only by permutation are grouped into six
composition families,
\[
MM,\quad
MD,\quad
MU,\quad
DD,\quad
DU,\quad
UU.
\]
For example, $MD$ contains both $(M,D)$ and $(D,M)$.
The downstream rules used here are symmetric, so this grouping does not alter
their target values.

The four replacement rules are
\[
\mathrm{ANY\_MATCH},
\qquad
\mathrm{ANY\_MISMATCH},
\qquad
\mathrm{ANY\_UNKNOWN},
\qquad
\mathrm{HOMOGENEOUS}.
\]
For
$s_1,s_2\in\mathcal S$,
\begin{align*}
\phi_{\mathrm{AM}}(s_1,s_2)
&=
\mathbf 1\{M\in\{s_1,s_2\}\},
\\
\phi_{\mathrm{AD}}(s_1,s_2)
&=
\mathbf 1\{D\in\{s_1,s_2\}\},
\\
\phi_{\mathrm{AU}}(s_1,s_2)
&=
\mathbf 1\{U\in\{s_1,s_2\}\},
\\
\phi_{\mathrm H}(s_1,s_2)
&=
\mathbf 1\{s_1=s_2\}.
\end{align*}
Thus the candidate-state mapping remains fixed while the downstream
composition rule changes, directly instantiating
$\mathcal H_S^{(2)}$.

\paragraph{Leave-one-composition-out construction.}
Each fold withholds one composition family from source learning and downstream
readout adaptation.
The six folds respectively hold out
\[
MM,\quad MD,\quad MU,\quad DD,\quad DU,\quad UU.
\]

For the representation-independent oracle audit, a state pair
$(z_1,z_2)\in\mathcal S^2$
is represented by the six-dimensional semantic interface
\begin{equation}
\psi(z_1,z_2)
=
\operatorname{concat}
\left(
e(z_1)+e(z_2),
\;
e(z_1)\odot e(z_2)
\right),
\label{eq:hard_semantic_interface}
\end{equation}
where
$e(z)\in\{0,1\}^3$
is the one-hot encoding of the candidate state.

\paragraph{Oracle ridge readout.}
For one replacement rule and one held-out composition, let
\[
\{(\psi_i,y_i)\}_{i=1}^{n},
\qquad
n=5,
\]
denote the five non-held-out oracle composition prototypes and their binary
targets.
Targets are encoded as
\[
t_i
=
2y_i-1
\in
\{-1,+1\}.
\]
The oracle evaluator fits
\begin{equation}
(\widehat w,\widehat b)
=
\arg\min_{w,b}
\left\{
\frac{1}{n}
\sum_{i=1}^{n}
\left(
t_i-w^\top\psi_i-b
\right)^2
+
\lambda_{\mathrm{ridge}}
\|w\|_2^2
\right\},
\qquad
\lambda_{\mathrm{ridge}}=10^{-3}.
\label{eq:oracle_ridge_readout}
\end{equation}
The intercept $\widehat b$ is not regularized.
The semantic feature vectors in
Eq.~(\ref{eq:hard_semantic_interface})
are used directly without per-coordinate centering or standardization;
``normalized'' refers to the $1/n$ normalization of the squared-error
objective, equivalently of the Gram and right-hand-side terms.
For the held-out prototype $\psi_\ast$, the binary prediction is
\begin{equation}
\widehat y_\ast
=
\mathbf 1
\left\{
\widehat w^\top\psi_\ast+\widehat b
\ge0
\right\}.
\label{eq:oracle_ridge_decision}
\end{equation}

This leave-one-composition-out calculation determines whether a family--rule
cell lies within the extrapolation capability of the specified downstream
readout when the true candidate states are available.
It is therefore a property of the oracle state interface and fixed readout
protocol, not of any learned representation.

Crossing six held-out composition families with four replacement rules gives
\[
6\times4=24
\]
candidate cells.
The oracle-solvability criterion retains
\[
21
\]
of them.

This $24\rightarrow21$ restriction is not a generic
two-class-support filter.
The target on an individual held-out composition prototype may be constant.
The relevant question is whether a readout fitted on the five non-held-out
oracle prototypes extrapolates the correct target to the held-out prototype.


\subsection{Frozen-Representation Reuse Protocol}
\label{app:frozen_reuse_protocol}

All reuse experiments separate source learning from downstream adaptation.
After source learning,

\begin{enumerate}
    \item the source-task head is discarded;
    \item the encoder is frozen;
    \item replacement-task gradients do not update the encoder; and
    \item a new downstream rule is fitted using target-task supervision.
\end{enumerate}

For the primary held-out-family analysis, five independently trained
representations are compared under matched source-training seeds.

Target supervision varies over
\[
K\in\{1,2,4,8,16\}
\]
examples per state together with a full-support condition.
Adaptation measurements are paired across compared representation conditions
according to the pre-specified evaluator.

The auxiliary structural readout used during source learning is discarded and
is never supplied to the downstream reuse evaluator.


\subsection{Source-Matched Novel-Rule Evaluation}
\label{app:source_matched_protocol}

The Shapes3D experiment in
Section~\ref{sec:task_success_reuse}
tests whether identical source-task performance determines reuse.

The pre-specified candidate registry contains
\[
2032
\]
rule--evaluation cells.
A representation-independent audit partitions these into
\[
1008\ \text{oracle-solvable},
\qquad
1008\ \text{oracle-unsolvable},
\qquad
16\ \text{single-class}
\]
cells.
Novelty and admissibility screening retains
\[
1004
\]
of the oracle-solvable cells for the source-matched analysis.

All five source-training seeds under C0, C1, and C4 attain source validation
accuracy
\[
1.0000,
\]
yielding 15 exactly source-matched models.

Registry inclusion is determined independently of the measured reuse
performance of the learned representations.


\subsection{Representation-Independent Admissibility Criteria}
\label{app:admissibility_protocols}

Admissibility is determined without using the performance of C0, C1, C4, or
any other learned representation.

\paragraph{Source-matched novel-rule protocol.}
Candidate cells are screened by the representation-independent oracle,
novelty, and registry criteria described above.
The resulting primary registry contains
\[
1004
\]
cells.

\paragraph{Held-out candidate-state protocol.}
For each of the 24 family--rule cells, candidate states are converted to the
semantic interface in
Eq.~(\ref{eq:hard_semantic_interface}).
The oracle ridge readout in
Eq.~(\ref{eq:oracle_ridge_readout})
is fitted using the five non-held-out composition prototypes and evaluated on
the sixth using
Eq.~(\ref{eq:oracle_ridge_decision}).
A cell is designated
\emph{oracle-solvable}
when this prediction equals the registered target on the held-out composition.

The resulting disposition is:

\begin{table}[t]
\centering
\small
\setlength{\tabcolsep}{4pt}
\caption{
Leave-one-composition-out oracle disposition of the 24 candidate
family--rule cells.
``Included'' means that the fixed downstream readout extrapolates the
replacement rule correctly from the five non-held-out oracle composition
prototypes to the held-out prototype.
}
\label{tab:oracle_loo_registry}
\begin{tabular}{lcccc}
\hline
Held-out family
& ANY\_MATCH
& ANY\_MISMATCH
& ANY\_UNKNOWN
& HOMOGENEOUS
\\
\hline
$MM$ & Included & Included & Included & Excluded \\
$MD$ & Included & Included & Included & Included \\
$MU$ & Included & Included & Included & Included \\
$DD$ & Included & Included & Included & Excluded \\
$DU$ & Included & Included & Included & Included \\
$UU$ & Included & Included & Included & Excluded \\
\hline
\end{tabular}
\end{table}

The three excluded cells are
\[
MM\times\mathrm{HOMOGENEOUS},
\qquad
DD\times\mathrm{HOMOGENEOUS},
\qquad
UU\times\mathrm{HOMOGENEOUS}.
\]
For each of these cells, the held-out target is $1$, whereas the
leave-one-composition-out oracle readout predicts $0$.

When all six oracle composition prototypes are simultaneously available, all
four replacement rules are realizable by the same oracle interface and
readout family.
The three exclusions are therefore specific to leave-one-composition-out
readout extrapolation.
They are neither single-class exclusions nor evidence that a learned
representation has lost the corresponding candidate state.

The same 21-cell registry is used for every source-learning condition,
representation seed, target-supervision level, and paired comparison.


\subsection{Aligned State-Recoverability Diagnostic}
\label{app:aligned_state_f1}

Section~\ref{sec:reuse_mechanism}
uses state macro-F1 as a diagnostic of candidate-state recoverability from the
frozen representation.

Let $\Pi_3$ denote the $3!$ permutations of the three state labels.
If $\widehat s(x)$ is a diagnostic state prediction, define
\[
\mathrm{F1}_{\mathrm{aligned}}
=
\max_{\pi\in\Pi_3}
\mathrm{MacroF1}
\left(
\pi(\widehat s(x)),
s(x)
\right).
\]

The global permutation removes arbitrary naming differences among the three
states.
It does not modify the frozen representation and is not supplied to the
downstream reuse evaluator.
The metric is therefore a diagnostic of whether the three-state structure is
recoverable up to one global relabeling, not the reuse score itself.


\subsection{Coordinate-Intervention and Rescue Protocol}
\label{app:coordinate_intervention_protocol}

Coordinate interventions are applied after representation learning to the same
frozen representation.
Each transformation is bijective, so the original representation remains
exactly recoverable when the applicable convention is known.

The intervention conditions are:

\begin{itemize}
    \item \textbf{Identity}: no coordinate transformation;
    \item \textbf{Global}: one reversible transformation shared by all samples;
    \item \textbf{2-way}: samples use one of two reversible conventions;
    \item \textbf{4-way}: samples use one of four conventions;
    \item \textbf{8-way}: samples use one of eight conventions;
    \item \textbf{Sample-wise}: convention identity varies at the sample level.
\end{itemize}

The blind downstream evaluator receives the transformed representation but not
the applicable convention identity.
The multi-convention conditions therefore increase the heterogeneity of
sample-dependent coordinate conventions while withholding the information
needed to invert them.

These finite-supervision interventions provide an empirical diagnostic of
shared interpretability. They do not correspond one-to-one to the exact
coordinate-conflict set
$\mathcal K_{U,h}^{\mathcal G}$.

For the rescue condition, the convention identity is supplied and the exact
inverse transformation is applied before downstream decoding.
The reconstruction audit gives
\[
\max_i
\|
\widehat z_i-z_i
\|
=
0,
\]
and the convention-aware evaluator returns to Identity-level reuse.


\subsection{Prospective Conditional-Accessibility Protocol}
\label{app:conditional_accessibility_protocol}

A separate CIFAR-100 protocol varies task-relevant conditional distinctions
during source learning and evaluates frozen-representation reuse across the
resulting intervention levels.

Let $C$ denote the CIFAR-100 coarse class and $F$ the corresponding fine
class.
Each coarse class contains five fine classes.
For intervention level
\[
q\in\{1,2,3,4,5\},
\]
the source-learning protocol uses a deterministic auxiliary target
\begin{equation}
A_q
=
m_q(C,F).
\label{eq:accessibility_mapping}
\end{equation}
The mappings $m_q$ are fixed before representation training and remain the same
across representation seeds and across the two tested encoder architectures.

\paragraph{Five-state partition construction.}
Within each coarse class, the five fine states are grouped by the auxiliary
mapping.
Under the balanced five-state calibration distribution, the five intervention
levels have the following block-size profiles:
\[
\begin{array}{ccl}
Q1 &: & (5),\\
Q2 &: & (3,2),\\
Q3 &: & (2,2,1),\\
Q4 &: & (2,1,1,1),\\
Q5 &: & (1,1,1,1,1).
\end{array}
\]
Thus Q1 assigns all five fine states to one auxiliary value, whereas Q5
distinguishes all five.
The intermediate conditions have progressively lower calibrated residual
fine-state uncertainty, but this block-size ordering alone is not used to
assert a nested-refinement relation between their identity-level partitions.

For each coarse class $c$, let
\[
\mathcal F_c
=
\{f_{c,1},\ldots,f_{c,5}\}
\]
denote its five fine classes.
The actual experimental auxiliary target is determined by the fixed lookup
\[
m_q(c,f_{c,j}),
\qquad
q=1,\ldots,5,\quad j=1,\ldots,5.
\]
The identity-level assignments are frozen before representation training and
are identical across representation seeds and encoder architectures.
The anonymous protocol configuration records these lookup tables explicitly.
The information-theoretic calibration below depends on the induced partition
sizes; the learned-representation analysis uses the actual frozen lookup
rather than only the block-size profile.

\paragraph{Information-theoretic calibration.}
Let
\[
H_q
=
H_{\mathrm{cal}}
\left(
F\mid C,A_q
\right)
\]
denote the conditional entropy under the fixed balanced calibration
distribution.
For a partition $\mathcal P_q$ of the five fine states,
\[
H_q
=
\frac{1}{5}
\sum_{B\in\mathcal P_q}
|B|\log_2|B|.
\]
The Q1 baseline is
\[
H_1
=
\log_2 5
=
2.3219280949
\quad\text{bits}.
\]

We define the normalized conditional-accessibility coordinate
\begin{equation}
a_q
=
1-
\frac{
H_{\mathrm{cal}}(F\mid C,A_q)
}{
H_1
}.
\label{eq:conditional_accessibility_fraction}
\end{equation}

The resulting intervention levels are:

\begin{table}[t]
\centering
\small
\caption{
Conditional-accessibility calibration.
The quantities are determined by the predefined auxiliary-target construction,
not by learned representation performance.
}
\label{tab:conditional_accessibility_levels}
\begin{tabular}{lccc}
\hline
Level
& $H_{\mathrm{cal}}(F\mid C,A_q)$
& $a_q$
& Oracle fine-state Bayes accuracy
\\
\hline
Q1 & $2.3219281$ & $0.0000000$ & $0.2$ \\
Q2 & $1.3509775$ & $0.4181657$ & $0.4$ \\
Q3 & $0.8000000$ & $0.6554588$ & $0.6$ \\
Q4 & $0.4000000$ & $0.8277294$ & $0.8$ \\
Q5 & $0.0000000$ & $1.0000000$ & $1.0$ \\
\hline
\end{tabular}
\end{table}

The intervention level is therefore a property of the predefined auxiliary
target, not a post-hoc estimate of how much fine-state information the learned
representation actually retains.

\paragraph{Where the intervention acts.}
The variable $A_q$ is used only as an auxiliary source-training target.
It is not concatenated to the frozen representation and is not supplied to the
downstream readout.

For a fixed architecture and matched seed identifier, Q1--Q5 correspond to
five separately trained representations,
\[
U_{s,1},\ldots,U_{s,5}.
\]
The five runs use the same source-task protocol but differ in the predefined
auxiliary target.
After source learning, the auxiliary head is discarded, the encoder is frozen,
and held-out future-rule reuse is evaluated under a common downstream
protocol.

The Q1--Q5 curve therefore compares representations produced by different
predefined source-learning interventions.
It is not obtained by progressively revealing side information to a single
frozen encoder.

\paragraph{Endpoint interpretation.}
The value
\[
a_1=0
\]
means that Q1 provides no reduction in the calibrated residual uncertainty
about $F$ relative to the Q1 baseline.
It does not mean that the learned representation contains no information useful
for downstream tasks.

Similarly,
\[
a_5=1
\]
means that the Q5 auxiliary construction fully resolves the controlled
fine-state component under the calibration distribution.
It does not imply that every downstream distinction is perfectly represented.

\paragraph{Source matching and replication.}
For each architecture, source-task matching is evaluated separately for
\[
5\ \text{matched seed blocks}
\times
5\ \text{Q levels}
=
25
\]
trained representations.
The accessibility construction, future-task protocol, data split, and
source-matching rule remain fixed across Q levels.

The empirical claim is restricted to this predefined source-learning
intervention: under source matching, condition-level mean held-out reuse
increases across Q1--Q5 for both tested encoder architectures.
The experiment does not establish universal monotonicity for arbitrary
representations, auxiliary objectives, task families, or decoder classes.


\subsection{Statistical Estimands and Replication Units}
\label{app:statistical_protocol}

\paragraph{Primary held-out-family analysis.}
Let
$A_{s,C}(K)$
denote aggregate reuse accuracy for source-training seed $s$, condition $C$,
and target-supervision level $K$, evaluated over the fixed 21-cell registry.

Define
\[
A_{s,C}^{\mathrm{LS}}
=
\frac{1}{5}
\sum_{K\in\{1,2,4,8,16\}}
A_{s,C}(K).
\]
For a comparison condition $C$ against Source-only C0, define
\[
\delta_{s,C}^{\mathrm{LS}}
=
A_{s,C}^{\mathrm{LS}}
-
A_{s,\mathrm{C0}}^{\mathrm{LS}},
\]
and
\[
\Delta_{\mathrm{LS}}(C)
=
\frac{1}{S}
\sum_{s=1}^{S}
\delta_{s,C}^{\mathrm{LS}},
\qquad
S=5.
\]

Analogously,
\[
\delta_{s,C}^{\mathrm{full}}
=
A_{s,C}(\mathrm{Full})
-
A_{s,\mathrm{C0}}(\mathrm{Full}),
\]
and
\[
\Delta_{\mathrm{full}}(C)
=
\frac{1}{S}
\sum_{s=1}^{S}
\delta_{s,C}^{\mathrm{full}}.
\]
The supervision-localized contrast is
\[
\Gamma(C)
=
\Delta_{\mathrm{LS}}(C)
-
\Delta_{\mathrm{full}}(C).
\]

The independently trained source representation is the primary replication
unit.
Repeated task cells, adaptation draws, and supervision levels contribute to a
seed-level aggregate and are not counted as independent representation
replicates.

Confidence intervals for paired representation comparisons resample at the
representation-seed level, preserving all within-seed task, adaptation, and
supervision measurements as one block.

\paragraph{Conditional-accessibility analysis.}
For architecture $A$, matched seed block $s$, and intervention level $q$, let
\[
R_{A,s,q}
\]
denote the frozen held-out reuse aggregate of the separately trained
Q-specific representation.

The condition-level mean is
\[
\overline R_{A,q}
=
\frac{1}{5}
\sum_{s=1}^{5}
R_{A,s,q}.
\]
The endpoint contrast is
\[
\Delta_{5-1}(A)
=
\overline R_{A,5}
-
\overline R_{A,1},
\]
computed before display rounding.

For each matched seed block, the accessibility--reuse trend is summarized by
the ordinary least-squares slope
\begin{equation}
\widehat\beta_{A,s}
=
\frac{
\sum_{q=1}^{5}
(a_q-\overline a)
(R_{A,s,q}-\overline R_{A,s})
}{
\sum_{q=1}^{5}
(a_q-\overline a)^2
},
\label{eq:accessibility_seed_slope}
\end{equation}
where
\[
\overline a
=
\frac{1}{5}
\sum_{q=1}^{5}a_q,
\qquad
\overline R_{A,s}
=
\frac{1}{5}
\sum_{q=1}^{5}R_{A,s,q}.
\]
The architecture-level slope is
\[
\overline\beta_A
=
\frac{1}{5}
\sum_{s=1}^{5}
\widehat\beta_{A,s}.
\]

The five Q-specific models belonging to one seed identifier form one matched
block.
They are separately trained representations, but their Q1--Q5 trajectory is
kept together for inference.

For the reported bootstrap interval, complete seed blocks are resampled with
replacement.
With five seed blocks, exhaustive ordered resampling contains
\[
5^5
=
3125
\]
bootstrap samples.
For each resample, the architecture-level statistic is the mean of the
selected seed-block slopes.
The reported 95\% interval is the equal-tailed percentile interval of this
exhaustively enumerated empirical bootstrap distribution.
Here ``exact bootstrap'' refers to exact enumeration of the bootstrap
resampling space, not to exact finite-sample confidence coverage.

Two empirical statements are kept separate:

\begin{enumerate}
    \item the five condition-level means
    $\overline R_{A,1},\ldots,\overline R_{A,5}$
    are strictly ordered; and
    \item all five seed-block slopes
    $\widehat\beta_{A,s}$
    are positive.
\end{enumerate}

A positive seed-block slope does not imply that every adjacent Q transition is
positive within that block.


\subsection{Relationship Between Experimental Protocols}
\label{app:protocol_relationships}

The experiments address different parts of the claim--evidence chain, and
their absolute accuracies are not treated as directly interchangeable.

The Shapes3D source-matched novel-rule registry tests the empirical separation
\[
\text{source-task success}
\not\Rightarrow
\text{representation reuse}.
\]

The structural intervention tests whether preserving a candidate state is
accompanied by increased candidate-state recoverability and reuse on tasks
using that state.

The coordinate intervention instead manipulates the availability of a shared
coordinate convention while keeping the original frozen representation
recoverable through inversion.

The 21-cell held-out-family protocol tests the constructive case in which
anticipated tasks factor through the candidate state.
It is therefore the empirical counterpart of the candidate-state construction
in
Section~\ref{sec:structure_preserving_learning},
not a claim that SPR improves every replacement-task family.

Finally, the CIFAR-100 conditional-accessibility experiment varies a
predefined auxiliary source-learning intervention under source matching and
tests whether the resulting intervention--reuse relationship is reproduced
with two encoder architectures.
The calibrated coordinate $a_q$ characterizes the auxiliary source-learning
condition and is not a direct measurement of structural accessibility in the
trained representation.

Scientific comparisons are therefore made within protocol using paired or
pre-specified contrasts; absolute accuracies from these distinct protocols are
not compared as scientific contrasts.


\section{Supporting Experimental Results}
\label{app:supporting_results}


\subsection{Source-Matched Reuse Separation}
\label{app:claim1_results}

In the Shapes3D protocol, all five seeds under C0, C1, and C4
achieve source validation accuracy
\[
1.0000.
\]
Across the 1004 admissible novel-rule evaluations, mean low-shot reuse is
\[
0.6410
\quad\text{for C0},
\]
\[
0.6040
\quad\text{for C1},
\]
and
\[
0.6031
\quad\text{for C4}.
\]
The paired low-shot differences relative to C0 are
\[
-0.0370
\quad\text{for C1},
\qquad
-0.0379
\quad\text{for C4},
\]
with the same direction across all five paired source seeds.

Under full target support, the corresponding differences contract to
\[
-0.0028
\qquad\text{and}\qquad
-0.0040.
\]

Thus identical source-task validation accuracy does not determine
frozen-representation reuse, and the observed separation is substantially
larger under limited target supervision.

\begin{figure*}[t]
\centering
\includegraphics[
    width=\textwidth
]{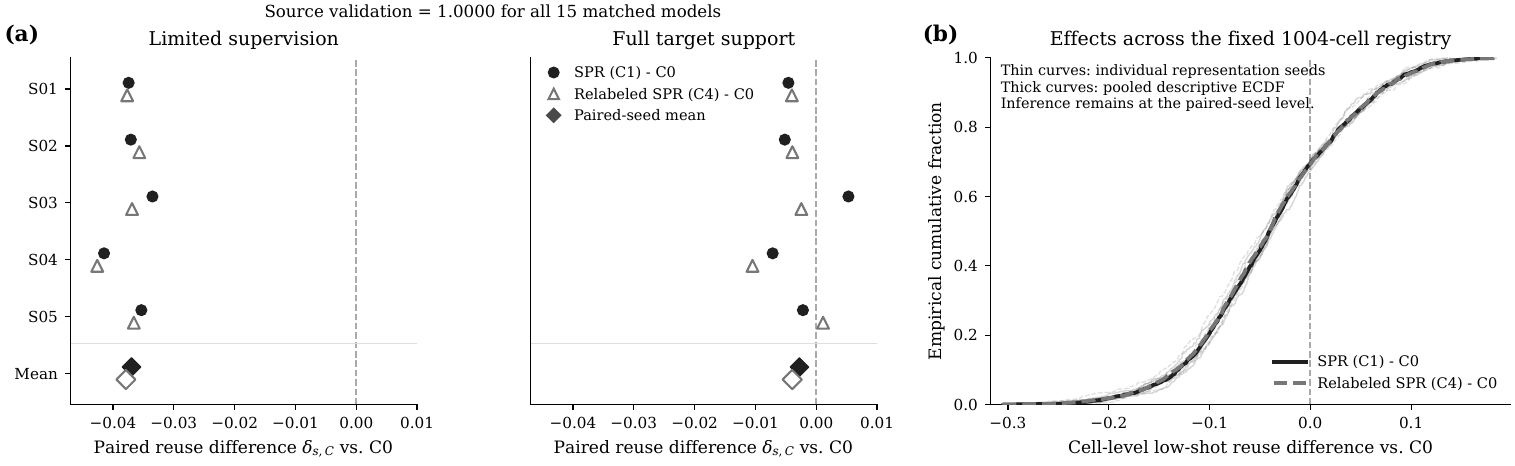}
\caption{
\textbf{Paired audit of source-matched reuse separation.}
All C0, C1, and C4 representations achieve identical source-task validation
accuracy (1.0000 across all 15 trained models).
\textbf{(a)}
Paired reuse differences relative to the corresponding C0 representation for
each independently trained source seed.
Under limited target supervision, C1 and C4 show the same direction of
separation from C0 across all five seeds, whereas the differences contract
substantially under full target support.
\textbf{(b)}
Descriptive distributions of cell-level low-shot paired effects across the
fixed 1004-cell admissible novel-rule registry.
Thin curves show individual representation seeds and thick curves show pooled
descriptive empirical cumulative distribution functions.
Registry cells expose heterogeneity across replacement tasks but are not
treated as independent statistical replicates; inference is performed at the
paired representation-seed level.
}
\label{fig:app_source_matched_reuse_audit}
\end{figure*}

Figure~\ref{fig:app_source_matched_reuse_audit} exposes the corresponding
seed-level replication structure and the heterogeneity across the fixed
admissible novel-rule registry.

\subsection{Candidate-State Recoverability and Structural Controls}
\label{app:structural_results}

Across five matched seeds, aligned held-out state macro-F1 changes from
\[
0.5547\pm0.1807
\]
under C0 to
\[
0.9646\pm0.0052
\]
under C1.
The paired change is
\[
+0.4099,
\qquad
95\%\ \mathrm{CI}
=
[0.2719,\;0.5479].
\]

Low-shot reuse changes from
\[
0.6491\pm0.1130
\]
to
\[
0.8356\pm0.0274,
\]
with paired difference
\[
+0.1866,
\qquad
95\%\ \mathrm{CI}
=
[0.1012,\;0.2875].
\]
The reuse contrast is positive for all five paired seeds.

C4 retains similar state recoverability:
\[
0.9572\pm0.0109
\]
versus
\[
0.9646\pm0.0052
\]
for C1.
The direct C1--C4 reuse contrast is
\[
-0.0072,
\qquad
95\%\ \mathrm{CI}
=
[-0.0375,\;0.0203].
\]

Random auxiliary and Shuffled-state controls do not reproduce the C1 reuse
level.
The paired C1 advantages are
\[
+0.5845
\]
and
\[
+0.6576,
\]
respectively, with the same direction across all five matched seeds.

Within this protocol, task-aligned candidate-state preservation is associated
with higher recoverability and reuse.
This association does not establish a general monotone relationship between
arbitrary approximate state recovery and restricted reuse.

\begin{figure*}[t]
\centering
\includegraphics[
    width=\textwidth
]{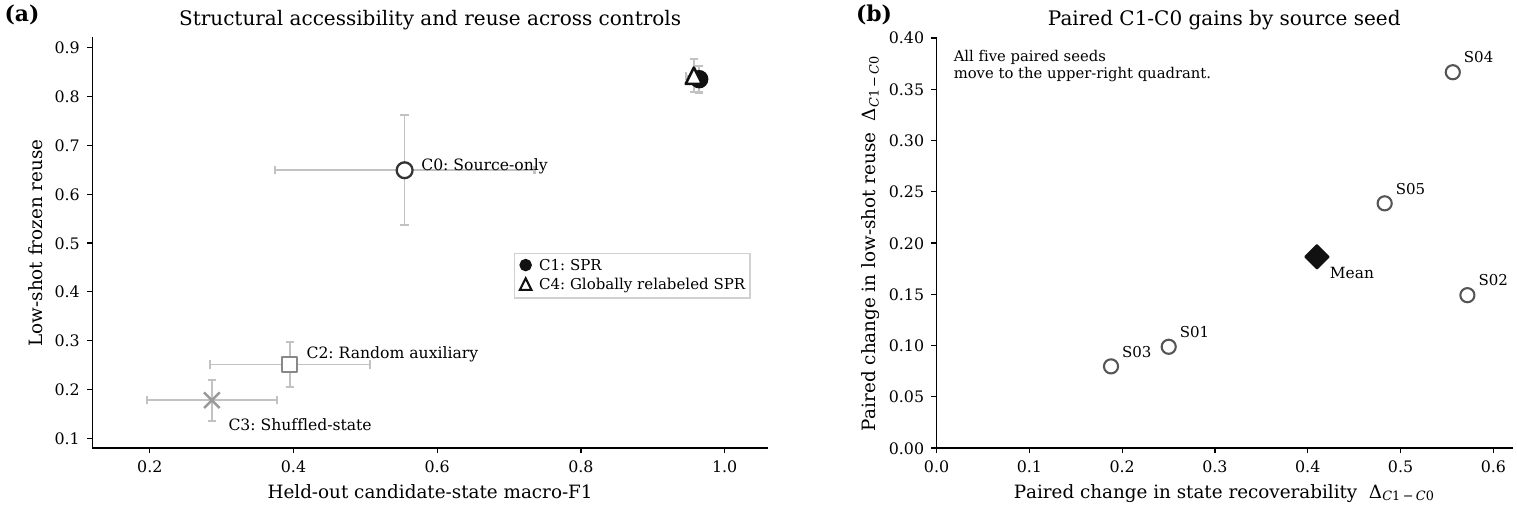}
\caption{
\textbf{Structural accessibility and low-shot reuse under matched controls.}
\textbf{(a)}
Condition-level relationship between held-out candidate-state recoverability
and low-shot frozen reuse across Source-only (C0), SPR (C1), Random auxiliary
(C2), Shuffled-state (C3), and Globally relabeled SPR (C4).
Markers show condition means and error bars show standard deviations across
five independently trained source seeds.
Improving candidate-state recoverability is accompanied by improved low-shot
reuse; auxiliary training without task-aligned state structure (C2) or with
broken observation--state correspondence (C3) does not reproduce the effect,
whereas a globally shared relabeling of the preserved state space (C4) remains
close to C1.
\textbf{(b)}
Paired C1--C0 gains for each source seed in held-out candidate-state
recoverability and low-shot reuse.
All five paired seeds lie in the upper-right quadrant; both gains are positive
for every source seed.
}
\label{fig:app_structural_accessibility_reuse}
\end{figure*}

Figure~\ref{fig:app_structural_accessibility_reuse} summarizes the
condition-level relationship between held-out candidate-state recoverability
and low-shot frozen reuse, and shows the paired C1--C0 gains for each
independently trained source seed.

\subsection{Coordinate-Convention Intervention and Rescue}
\label{app:coordinate_results}

The same frozen representation is evaluated after reversible coordinate
interventions.

Mean low-shot reuse is
\[
0.4096
\]
under Identity and
\[
0.4099
\]
under one globally shared transformation.
Under increasingly heterogeneous sample-dependent conventions, reuse is
\[
0.3449,\quad
0.2679,\quad
0.1836,\quad
0.0154
\]
for 2-way, 4-way, 8-way, and sample-wise conditions, respectively.

The ordering is reproduced for each of the three source-training seeds, and
the Identity-to-sample-wise difference is
\[
0.3942.
\]

When convention identity is supplied and the corresponding inverse is
applied, reuse returns to
\[
0.4096
\]
in every condition.
The reconstruction audit gives
\[
\max_i
\|
\widehat z_i-z_i
\|
=
0.
\]

Within this finite-supervision intervention, the degradation is associated
with unavailable sample-specific coordinate conventions; convention-aware
inversion exactly recovers the common-coordinate representation.
This intervention is not a direct numerical measurement of the exact
coordinate-conflict set in Theorem~1.

\begin{figure*}[t]
\centering
\includegraphics[
    width=\textwidth
]{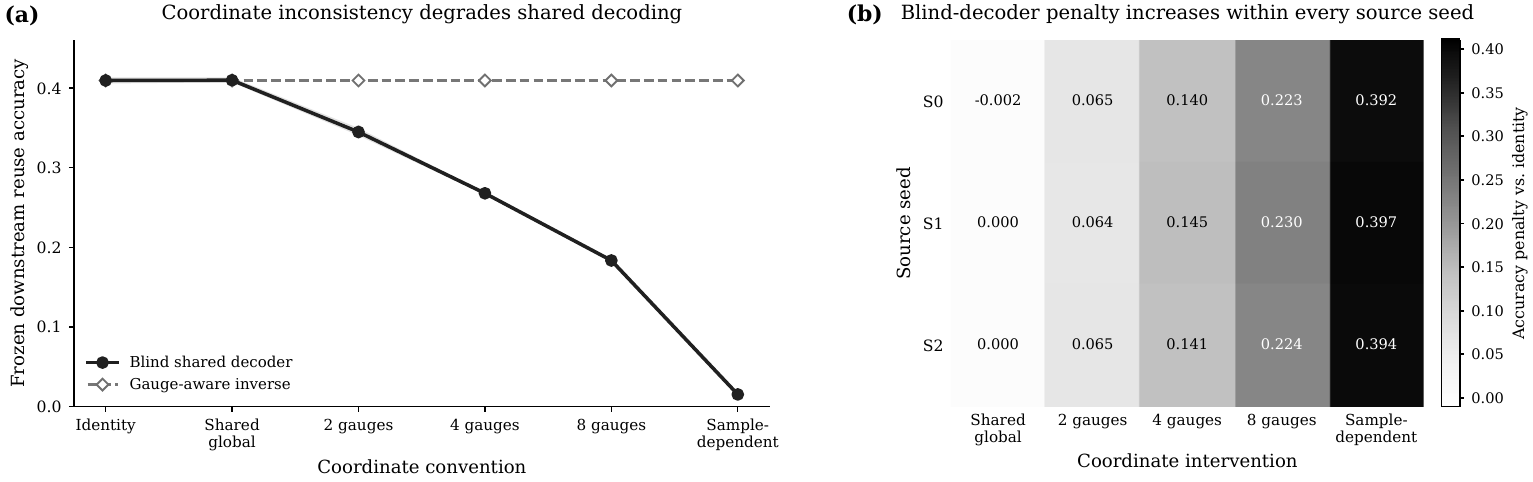}
\caption{
\textbf{Coordinate inconsistency degrades shared decoding and is exactly
rescued by convention-aware inversion.}
\textbf{(a)}
Frozen downstream reuse under increasingly heterogeneous reversible coordinate
conventions.
A single globally shared reparameterization leaves blind reuse essentially
unchanged relative to the identity convention, whereas reuse decreases
progressively as the number of unresolved conventions increases, reaching its
minimum under sample-dependent conventions.
Supplying the applicable inverse convention restores the identity-level reuse
accuracy for every intervention condition; the frozen reconstruction audit
reports zero residual relative to the common-coordinate representation.
\textbf{(b)}
Blind-decoder accuracy penalty relative to the identity convention for each
independently trained source representation.
For every source seed, the penalty increases from the shared-global control
through the 2-, 4-, and 8-convention interventions to the sample-dependent
condition.
}
\label{fig:app_coordinate_consistency_rescue}
\end{figure*}

Figure~\ref{fig:app_coordinate_consistency_rescue} summarizes the
coordinate-consistency continuum and the corresponding gauge-aware rescue,
and exposes the blind-decoder penalty separately for each independently
trained source representation.

\subsection{Reuse Across Held-Out Candidate-State Task Families}
\label{app:heldout_family_results}

The primary compatible-family analysis uses the fixed 21-cell admissible
registry and
\[
K\in\{1,2,4,8,16\}
\]
for the limited-supervision regime.

For C1 relative to C0,
\[
\Delta_{\mathrm{LS}}
=
+0.1715,
\qquad
95\%\ \mathrm{CI}
=
[0.0986,\;0.2591],
\]
whereas
\[
\Delta_{\mathrm{full}}
=
+0.0241,
\qquad
95\%\ \mathrm{CI}
=
[-0.0260,\;0.0992].
\]
Hence
\[
\Gamma
=
+0.1474,
\qquad
95\%\ \mathrm{CI}
=
[0.1044,\;0.1905].
\]

For C4 relative to C0,
\[
\Delta_{\mathrm{LS}}
=
+0.1771,
\qquad
95\%\ \mathrm{CI}
=
[0.1172,\;0.2593],
\]
\[
\Delta_{\mathrm{full}}
=
+0.0300,
\qquad
95\%\ \mathrm{CI}
=
[-0.0201,\;0.0990],
\]
and
\[
\Gamma
=
+0.1471,
\qquad
95\%\ \mathrm{CI}
=
[0.1193,\;0.1749].
\]

\begin{table}[t]
\centering
\small
\caption{
Paired held-out-family reuse effects relative to Source-only.
The inferential unit is the matched representation-training seed.
}
\label{tab:app_heldout_family}
\begin{tabular}{lccc}
\hline
Condition
& $\Delta_{\mathrm{LS}}$
& $\Delta_{\mathrm{full}}$
& $\Gamma$
\\
\hline
SPR
& $+0.1715$
& $+0.0241$
& $+0.1474$
\\
95\% CI
& $[0.0986,\,0.2591]$
& $[-0.0260,\,0.0992]$
& $[0.1044,\,0.1905]$
\\
\hline
Globally relabeled SPR
& $+0.1771$
& $+0.0300$
& $+0.1471$
\\
95\% CI
& $[0.1172,\,0.2593]$
& $[-0.0201,\,0.0990]$
& $[0.1193,\,0.1749]$
\\
\hline
\end{tabular}
\end{table}

The direct C1--C4 limited-supervision contrast is
\[
-0.0056,
\qquad
95\%\ \mathrm{CI}
=
[-0.0284,\;0.0173],
\]
with no consistent direction across seeds.

The positive $\Gamma$ contrasts indicate larger reuse gains under limited
target supervision. The full-support effects for C1 and C4 remain unresolved
because both confidence intervals include zero.

\begin{figure*}[t]
\centering
\includegraphics[
    width=\textwidth
]{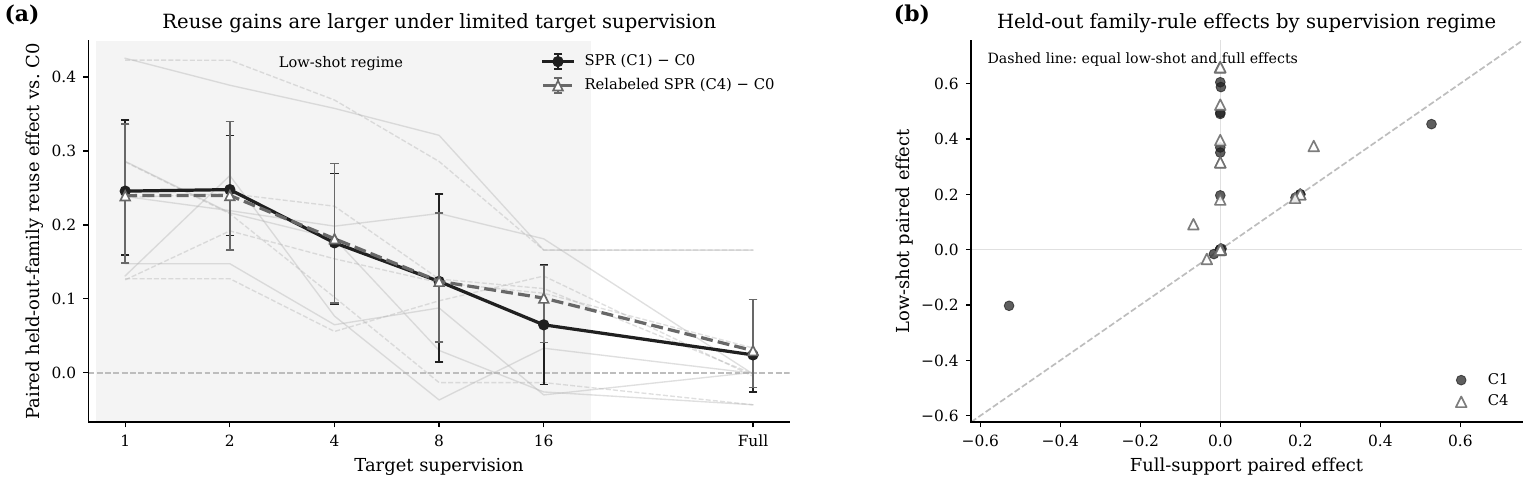}
\caption{
\textbf{Held-out-family reuse gains are concentrated under limited target
supervision.}
\textbf{(a)}
Paired reuse effects relative to Source-only across target-supervision levels
$K\in\{1,2,4,8,16\}$ and the full-support condition.
Solid circles denote SPR (C1) and open triangles denote Globally relabeled SPR
(C4); thick curves show paired means across the five representation-training
seeds, error bars show bootstrap 95\% confidence intervals, and faint curves
show the corresponding seed-level paired effects.
For both conditions, paired reuse gains are larger at each predefined
limited-supervision level than under full target support.
\textbf{(b)}
Descriptive low-shot and full-support paired effects for the 21 admissible
held-out family--rule evaluations.
Points above the diagonal have larger low-shot than full-support effects.
The cell-level distribution shows heterogeneity across replacement tasks and
is descriptive; inferential comparisons remain defined at the paired
representation-training-seed level.
}
\label{fig:app_heldout_family_lowshot_gains}
\end{figure*}

Figure~\ref{fig:app_heldout_family_lowshot_gains} reports the corresponding
supervision profile and the variation of paired effects across the admissible
held-out family--rule evaluations.

\subsection{Evidence Beyond the Primary Controlled Setting}
\label{app:generality_results}

\paragraph{Natural-image source-matched evidence.}
Across nine matched representation pairs on CIFAR-100, the mean absolute
within-pair source-performance gap is
\[
0.0036,
\]
while the low-shot SPR--Source-only reuse difference is
\[
+0.1718.
\]

\paragraph{Alternative source-learning formulation.}
Under ordinary supervised classification, source accuracy is
\[
0.7535
\]
for Source-only and
\[
0.7602
\]
for SPR.
After freezing the representations, the held-out-rule reuse difference is
\[
+0.0768
\]
in favor of SPR.

The source-success/reuse separation is also observed beyond the primary
relational construction.
This evidence does not establish universal superiority of one source objective
across arbitrary future-task families.


\subsection{Prospective Natural-Image Accessibility Replications}
\label{app:conditional_accessibility_results}

The CIFAR-100 experiments vary the predefined conditional-accessibility
source-learning intervention while enforcing source-task matching.
For each architecture, a matched seed block contains five separately trained
Q-specific representations.
Q1--Q5 are not different amounts of side information supplied to one fixed
encoder.

\begin{table}[t]
\centering
\small
\setlength{\tabcolsep}{3.4pt}
\caption{
Condition-level mean held-out reuse in the conditional-accessibility
replications.
Q5--Q1 is computed from the unrounded condition means.
}
\label{tab:app_conditional_accessibility}
\begin{tabular}{lrrrrrr}
\hline
Encoder
& Q1 & Q2 & Q3 & Q4 & Q5 & Q5--Q1
\\
\hline
ResNet-18
& 0.2670415
& 0.2916385
& 0.3000700
& 0.3075740
& 0.3212570
& 0.0542155
\\
DenseNet-121
& 0.2810590
& 0.3001825
& 0.3080730
& 0.3216465
& 0.3260565
& 0.0449975
\\
\hline
\end{tabular}
\end{table}

For both encoders, the condition-level means satisfy
\[
Q1<Q2<Q3<Q4<Q5.
\]
This statement concerns means across the five matched seed blocks and does not
assert strict adjacent monotonicity within every individual seed block.

\paragraph{ResNet-18.}
The five seed-level intervention--reuse slopes are
\[
0.0474311,\quad
0.0533076,\quad
0.0473905,\quad
0.0529626,\quad
0.0569875.
\]
All five are positive.
Their mean is
\[
0.0516159,
\]
with exhaustive seed-block bootstrap 95\% interval
\[
[0.0485212,\;0.0547106].
\]
All
\[
25/25
\]
Q-specific source-training runs satisfy the pre-specified source-matching
criterion.

\paragraph{DenseNet-121.}
The five seed-level intervention--reuse slopes are
\[
0.0425421,\quad
0.0456553,\quad
0.0467809,\quad
0.0490245,\quad
0.0458580.
\]
Again all five are positive.
Their mean is
\[
0.0459722,
\]
with exhaustive seed-block bootstrap 95\% interval
\[
[0.0440125,\;0.0477280].
\]
All
\[
25/25
\]
Q-specific DenseNet-121 runs satisfy the same pre-specified source-matching
criterion.

The monotone condition-level ordering is therefore an ordering across
predefined auxiliary source-learning interventions.
It should not be interpreted as obtaining five reuse points by progressively
revealing side information to one fixed encoder, nor as directly measuring the
amount of task-relevant information retained by each learned representation.

The positive intervention--reuse relationship is reproduced across the two
tested encoder architectures under the same CIFAR-100 hierarchy and
future-task construction. They do not establish architecture-independent
universality or a general theorem that increasing a source-side auxiliary
accessibility coordinate must monotonically improve restricted reuse.

\begin{figure*}[t]
\centering
\includegraphics[
    width=\textwidth
]{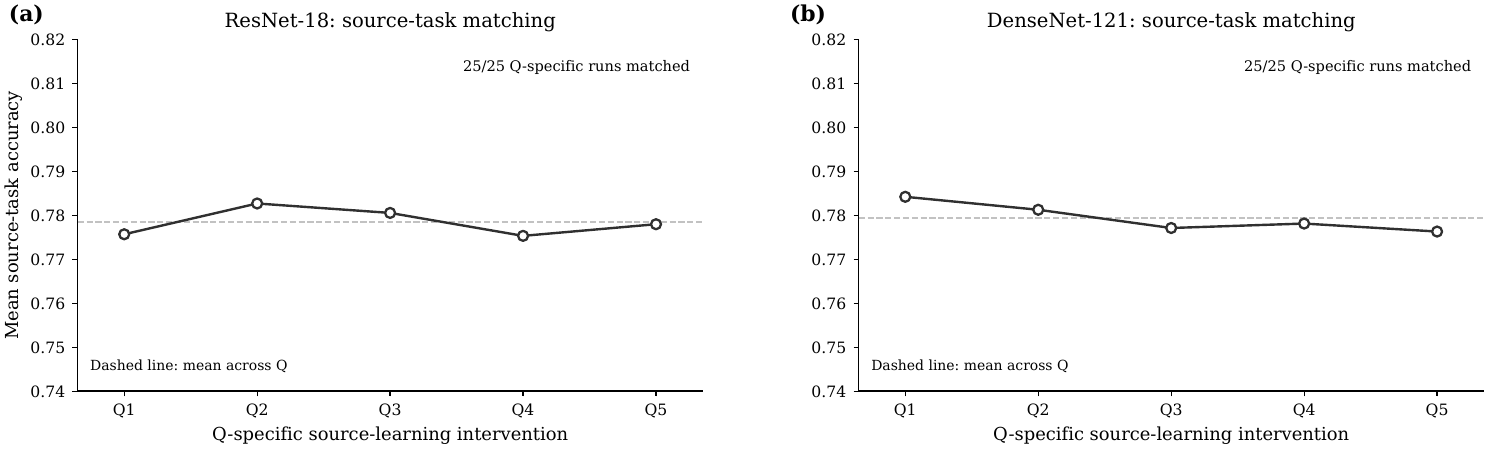}
\caption{
\textbf{Source-task matching across Q-specific natural-image replications.}
\textbf{(a)}
Mean source-task accuracy across the Q1--Q5 source-learning interventions for
the ResNet-18 replication.
Each Q level corresponds to a separately trained representation, and all
25 architecture--seed--Q source-training runs satisfy the pre-specified
source-matching criterion.
The dashed horizontal line denotes the mean source-task accuracy across the
five Q conditions.
\textbf{(b)}
The corresponding source-task matching audit for the DenseNet-121 replication.
All 25 Q-specific DenseNet-121 runs likewise satisfy the same pre-specified
matching criterion.
Together with Figure~\ref{fig:app_conditional_accessibility_replications},
this audit separates the observed intervention--reuse relationship from
uncontrolled failure of the source-task matching protocol; it does not require
source-task accuracy to be numerically identical across Q conditions.
}
\label{fig:app_source_matching_audit}
\end{figure*}

Figure~\ref{fig:app_source_matching_audit} complements the reuse results above
with a source-task matching audit across the Q1--Q5 interventions for both
encoder replications.

\begin{figure*}[t]
\centering
\includegraphics[
    width=\textwidth
]{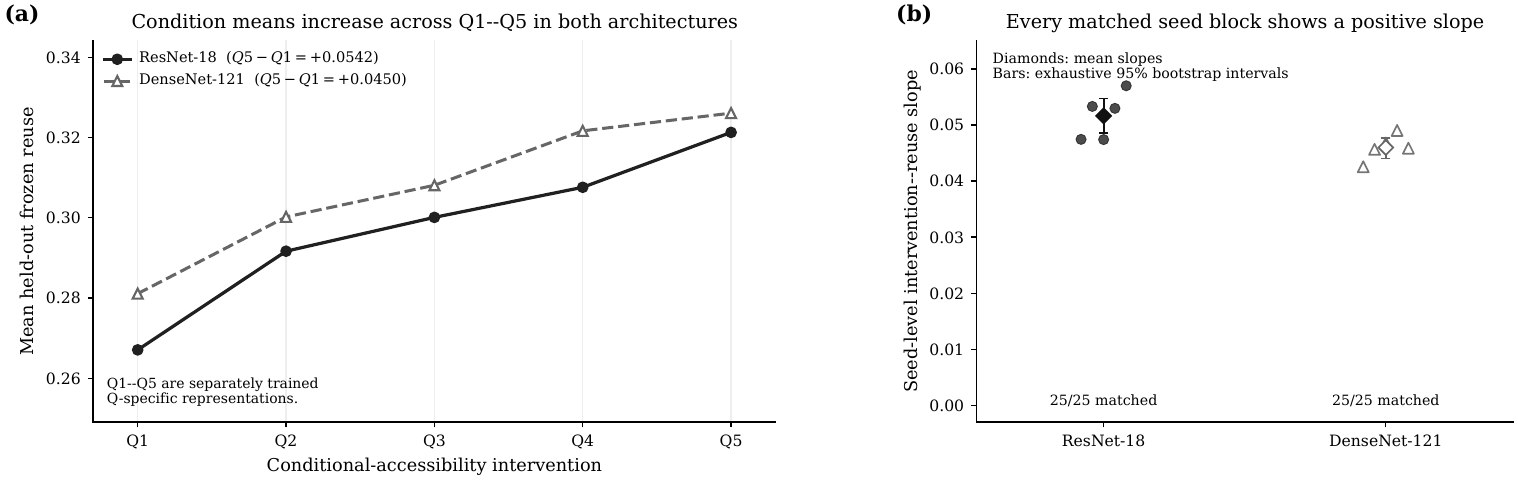}
\caption{
\textbf{Conditional-accessibility intervention--reuse relationship across two
encoder architectures.}
\textbf{(a)}
Held-out frozen reuse across the predefined Q1--Q5 source-learning
interventions for ResNet-18 and DenseNet-121.
Solid circles denote ResNet-18 and dashed open triangles denote DenseNet-121.
For each architecture, the legend reports the corresponding endpoint contrast
$Q5-Q1$.
Condition means increase from Q1 to Q5 for both architectures.
Q1--Q5 denote separately trained intervention-specific representations and
should not be interpreted as progressively revealing additional side
information to one fixed encoder.
\textbf{(b)}
Seed-level intervention--reuse slopes for the two architecture replications.
Individual markers denote matched seed blocks, diamonds denote mean slopes,
and vertical bars denote exhaustive seed-block bootstrap 95\% intervals.
All seed-level slopes are positive for both architectures.
All 25 Q-specific source-training runs for each architecture satisfy the
pre-specified source-matching criterion.
}
\label{fig:app_conditional_accessibility_replications}
\end{figure*}

Figure~\ref{fig:app_conditional_accessibility_replications} visualizes the
condition-level Q1--Q5 trajectories and the corresponding seed-level
intervention--reuse slopes for the two architecture replications.

\section{Implementation and Reproducibility Details}
\label{app:implementation_details}

This section records implementation properties required to interpret and
reproduce the claim-supporting experiments.
Protocol definitions and statistical estimands are given in
Appendix~\ref{app:experimental_protocols};
numerical results are given in
Appendix~\ref{app:supporting_results}.


\subsection{Datasets and Split Discipline}
\label{app:data_splits}

The primary controlled experiments use Shapes3D.
Natural-image extensions and the conditional-accessibility replications use
CIFAR-100.

Source learning and reuse evaluation are separated at both the model and
protocol levels.
The source model is trained before target adaptation begins.
The source head is subsequently removed and the encoder is frozen.
Target labels used to fit replacement rules never update the encoder.

In the held-out-composition benchmark, each evaluation fold withholds one of
the six composition families from source training.
The six held-out families are
\[
MM,\quad MD,\quad MU,\quad DD,\quad DU,\quad UU.
\]

The natural-image confirmatory experiments use fixed source, adaptation, and
evaluation partitions.
The conditional-accessibility replications use the same split construction and
future-task protocol across Q levels and across the two encoder architectures.
The official TEST partition is not used by these confirmatory evaluations.

Target-adaptation draws, task cells, and Q conditions are never counted as
additional independently trained representation replicates.


\subsection{Encoder and Source-Head Architectures}
\label{app:model_architecture}

Within each controlled comparison, the encoder architecture is fixed across
the source-learning conditions being compared.
Source-task and auxiliary heads are training components only and are removed
before reuse evaluation.

The ResNet replication uses a CIFAR-adapted ResNet-18 encoder
\citep{7780459} followed by a 128-dimensional representation projection.

The cross-architecture replication uses DenseNet-121
\citep{8099726} initialized without pretrained weights.
For CIFAR-sized inputs, its initial convolution is replaced by a
\[
3\rightarrow64
\]
convolution with kernel size $3$, stride $1$, padding $1$, and no bias, and
the initial pooling operation is removed.
The original classifier is removed and the backbone output is projected to
128 dimensions.

The two architecture replications use the same conditional-accessibility
construction, future-task protocol, representation dimension, and downstream
evaluation logic.
Q1--Q5 are not redefined for the second architecture.


\subsection{Source-Training Objectives and Matching}
\label{app:source_training_impl}

For the primary relational experiments, all compared representation
conditions optimize the same relational source objective.
C1 additionally optimizes the empirical structural loss,
\[
\mathcal L_{\mathrm{SPR}}
=
\widehat{\mathcal L}_{\mathrm{src}}
+
\lambda_{\mathrm{reuse}}
\widehat{\mathcal R}_{\mathrm{str}}.
\]
C4 uses the same structural objective after one fixed global bijection of
candidate-state labels.
C2 and C3 retain the auxiliary-training scaffold while replacing the
task-aligned structural target by random or shuffled targets.

In the exact Shapes3D comparison, all C0, C1, and C4 models attain source
validation accuracy
\[
1.0000.
\]

\paragraph{Conditional-accessibility source learning.}
The Q1--Q5 experiment trains a separate representation for each
architecture--seed--Q combination.
For level $q$, the auxiliary target is
\[
A_q=m_q(C,F).
\]
The auxiliary-loss coefficient is fixed at
\[
\lambda_{\mathrm{aux}}=1.0.
\]

The auxiliary target is used only during source representation learning.
It is neither concatenated to the representation nor supplied to the
downstream task.

For a matched seed block, all five Q runs share the same source-task,
optimization, split, and evaluation protocol and differ only in the predefined
Q-specific auxiliary target.

For ResNet-18, the five matched seed identifiers are
\[
101,\quad202,\quad303,\quad404,\quad505,
\]
and all 25 architecture--seed--Q runs satisfy the source-matching interval
\[
[0.77,\;0.79].
\]

For DenseNet-121, the five matched seed identifiers are
\[
1101,\quad1202,\quad1303,\quad1404,\quad1505,
\]
and all 25 runs satisfy the same source-matching criterion.


\subsection{Downstream Reuse Readout}
\label{app:downstream_readout_impl}

Every reported reuse evaluation operates on frozen encoder outputs.
The source head and training-time structural head are unavailable to the
downstream evaluator.

For limited-supervision experiments, the downstream rule is fitted only from
the registered target-adaptation observations for the corresponding budget.
The candidate-state task-family experiment evaluates
\[
K\in\{1,2,4,8,16\}
\]
examples per state and a full-support condition.

Adaptation indices are paired across representation conditions whenever a
paired model contrast is reported.
Thus C1--C0, C4--C0, and related paired effects use common evaluation tasks
and matched adaptation realizations.

Readout class, optimization, and regularization are fixed within each
evaluation protocol and are not tuned independently for each representation
condition.
Protocols using distinct evaluators are analyzed separately; their absolute
accuracies are not interpreted as measurements on one common scale.


\subsection{State-Recoverability Diagnostic}
\label{app:diagnostic_readout_impl}

The recoverability experiment fits a diagnostic state readout to the frozen
representation and reports macro-F1 after one global three-state label
alignment.
The alignment searches over the six permutations of
$\{M,D,U\}$
and selects a single global relabeling for the diagnostic metric.

This is permutation-invariant metric alignment.
It does not modify the encoder and is not supplied as side information to the
downstream reuse evaluator.

The same diagnostic is used for C0, C1, C4, Random auxiliary, and
Shuffled-state.


\subsection{Coordinate-Transformation Implementation}
\label{app:coordinate_impl}

All coordinate interventions are applied after representation learning to the
same frozen representation.

The transformation bank consists of reversible coordinate maps.
Identity applies no transformation; Global applies one transformation shared by
all samples; 2-way, 4-way, and 8-way use increasingly heterogeneous
sample-dependent convention sets; and the sample-wise condition varies the
convention at the sample level.
Convention identity is withheld from the blind downstream decoder.

For the rescue evaluation, convention identity is supplied and its exact
inverse is applied before downstream evaluation.
The reconstruction audit records
\[
\max_i
\|
\widehat z_i-z_i
\|
=
0,
\]
and reuse returns to
\[
0.4096
\]
in every rescued condition.

The rescue removes the imposed coordinate intervention; it does not retrain
the representation or provide additional task information.


\subsection{Task Registry and Admissibility Implementation}
\label{app:registry_impl}

All primary registry decisions are representation independent.

For the Shapes3D novel-rule analysis, the executable registry reproduces the
partition
\[
2032
=
1008
+
1008
+
16
\]
into oracle-solvable, oracle-unsolvable, and single-class candidate cells, and
the final novelty/admissibility criteria retain 1004 cells.

For the held-out candidate-state analysis, the executable oracle audit applies
the semantic interface in
Eq.~(\ref{eq:hard_semantic_interface}),
the ridge objective in
Eq.~(\ref{eq:oracle_ridge_readout}),
the zero-threshold binary rule in
Eq.~(\ref{eq:oracle_ridge_decision}),
and the leave-one-composition-out procedure defined in
Appendix~\ref{app:admissibility_protocols}.
It reproduces the 21 included and three excluded cells in
Table~\ref{tab:oracle_loo_registry}.

The resulting registries are fixed before learned-representation comparisons
and reused unchanged across source-learning conditions, seeds, and supervision
levels.


\subsection{Statistical Aggregation and Confidence Intervals}
\label{app:statistics_impl}

The primary inferential unit for representation comparisons is the
independently trained representation seed.

For the held-out candidate-state family, repeated target-adaptation
measurements are first aggregated within each
seed--condition--supervision cell.
The fixed admissible task registry is then aggregated to form
$A_{s,C}(K)$.
Limited-supervision effects, full-support effects, and $\Gamma$ are computed
from the seed-level paired estimands defined in
Appendix~\ref{app:statistical_protocol}.

Confidence intervals for paired representation comparisons resample complete
representation-seed blocks so that repeated task and adaptation measurements
from the same trained representation remain grouped.

For the conditional-accessibility experiment, one matched seed block contains
five separately trained Q-specific models.
The Q1--Q5 trajectory remains intact during resampling.
The seed-level trend coefficient is
Eq.~(\ref{eq:accessibility_seed_slope}),
and the architecture-level trend is the arithmetic mean of the five
seed-level coefficients.

Exact bootstrap inference enumerates the
\[
5^5=3125
\]
ordered resamples of the five matched seed blocks.
The reported 95\% interval is the equal-tailed percentile interval of the
resulting distribution of the mean trend coefficient.
Here ``exact'' means that the complete finite bootstrap resampling space is
enumerated; it does not denote an exact finite-sample coverage guarantee.

Condition-level monotonicity and positivity of the five seed-level slopes are
reported separately because they are logically distinct empirical properties.

\end{document}